\documentclass{article}

\usepackage{arxiv}

\usepackage[utf8]{inputenc} % allow utf-8 input
\usepackage[T1]{fontenc}    % use 8-bit T1 fonts
\usepackage[hidelinks]{hyperref}
\usepackage{amsmath}
\usepackage{url}            % simple URL typesetting
\usepackage{booktabs}       % professional-quality tables
\usepackage{amsfonts}       % blackboard math symbols
\usepackage{nicefrac}       % compact symbols for 1/2, etc.
\usepackage{microtype}      % microtypography
\usepackage{cleveref}       % smart cross-referencing
\usepackage{lipsum}         % Can be removed after putting your text content
\usepackage{graphicx}
\usepackage{natbib}
\usepackage{doi}
\usepackage{lineno}         % Added for line numbers

\title{Benchmarking Deep Learning Models for Dense Event Classification of Offshore Wind Infrastructure in Sentinel-1 Time Series}

\newif\ifuniqueAffiliation
\ifuniqueAffiliation % Standard variant of author block
\else
\usepackage{authblk}

\newbox{\orcid}\sbox{\orcid}{\includegraphics[scale=0.06]{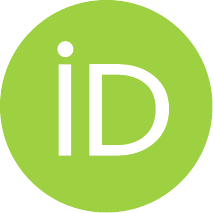}} 
\author[1]{%
	\href{https://orcid.org/0000-0002-7179-3664}{\usebox{\orcid}\hspace{1mm}Thorsten Hoeser}%
	\thanks{Corresponding author \texttt{thorsten.hoeser@dlr.de}}%
}

\author[1]{%
	\href{https://orcid.org/0000-0001-6181-0187}{\usebox{\orcid}\hspace{1mm}Felix Bachofer}%
}

\author[1,2]{%
	Claudia Kuenzer%
}

\affil[1]{Earth Observation Center (EOC), German Aerospace Center (DLR), Oberpfaffenhofen, 82234 Wessling, Germany}

\affil[2]{Institute for Geography and Geology, University of Wuerzburg, 97074 Wuerzburg, Germany}
\fi

\renewcommand{\headeright}{ }
\renewcommand{\undertitle}{Preprint}
\hypersetup{
pdftitle={Benchmarking Deep Learning Models for Dense Event Classification of Offshore Wind Infrastructure in Sentinel-1 Time Series},
pdfauthor={Thorsten~Hoeser, Felix~Bachofer, Claudia~Kuenzer},
}

\begin{document}
\maketitle
\begin{abstract}
Given the goals set by the three major markets China, the European Union, and the United Kingdom for extending offshore wind installed capacity over the upcoming 10 to 25 years, deployment activities for offshore wind turbines have to increase significantly. Monitoring of offshore wind energy infrastructure life cycles, especially during the deployment phase, is an important contribution for stakeholders to make informed decisions. Earth observation missions such as ESA's Sentinel-1 Synthetic Aperture Radar (SAR) mission produce large data archives that, when exploited at the single-acquisition level, enable the global monitoring of offshore wind infrastructure. Turning these high-volume archives into valuable information requires algorithms that automatically extract single event labels from dense time series at a global scale. In this study, we present a structured comparison of ten deep learning model--training variants for the dense classification of Sentinel-1 based offshore wind infrastructure time series, aiming to advance expert-driven, rule-based event classification of this task.
We trained LSTM, Transformer, and fully connected model variants with monotemporal, unidirectional, and bidirectional context awareness, each with and without self-supervised pretraining. Among these, the supervised BiLSTM performs best, raising the area under the collapsed edit similarity curve from 0.7853 for the rule-based baseline to 0.8509, and the perfect match rate from 0.3508 to 0.5063. Combining the BiLSTM predictions with the existing baseline labels in a label-transition-minimising ensemble yields the most user-friendly result, further improving agreement with the test data and recovering performance on under-represented classes.
Using these improved labels, we isolate the deployment phase of individual turbines at a global scale and conduct a regional and subregional analysis covering 2016-01-01 to 2025-03-31, reporting median deployment durations of 84~d (China), 242~d (EU), and 258~d (UK). Deployment-related drivers, including legal regulations such as subsidies, and environmental conditions, emerge clearly from the analysed results across multiple spatial scales.
\end{abstract}

% keywords can be removed

%\linenumbers                % Enable line numbering

\section{Introduction}
\label{sec:introduction}

The installed capacity of offshore wind infrastructure reached 91.378~GW as of February 2026, mainly distributed among the three major markets China (45.807~GW), the European Union (23.254~GW), and the United Kingdom (15.963~GW) \citep{gem_download_data}, with over 15,000 offshore wind turbines currently operational \citep{HOESER2026100451}. Official targets are to extend the installed capacity of offshore wind energy to 40~GW for the United Kingdom by 2030 \citep{UKGov2021} and to 300~GW for the European Union by 2050 \citep{EC2020}, while China announced an increase of solar and wind energy installed capacity to 3,600~GW, six times its 2020 levels, by 2035 \citep{UNClimateSummit2025}, which also includes offshore infrastructure. Reflecting these targets against the levels reached in 2025, major deployment activities will take place over the upcoming 10 to 25 years, developing new regions but also already established hot spots such as the North Sea Basin (EU and UK), and densifying offshore wind turbine installations along the Chinese coast.

With current high-capacity offshore wind turbine designs ranging, for example, from 15~MW (Vestas V236-15.0~MW \citep{enbw2025hedreiht}), commercially deployed in the German He Dreiht offshore wind farm, to 26~MW (Dongfang Electric DEW-26~MW-310), currently being tested \citep{sasac2025china26mw}, reaching the European target would require adding approximately 276.75~GW, corresponding to 18,450 15~MW or 10,644 26~MW offshore wind turbines. Compared to the 3,920 turbines present in the European Union in Q1~2025 \citep{HOESER2026100451}, this poses a considerable effort in infrastructure deployment.

Scaling up offshore wind energy infrastructure is therefore a major challenge in current markets. For each turbine, the most engineering- and logistically-challenging as well as impactful process is its deployment phase. Monitoring this phase across space and time will become increasingly important, as it provides a broader picture that goes beyond single projects and makes deployment characteristics comparable between regions and over time. Since offshore wind turbine (OWT) deployment lies at the interface of engineering, environmental, logistical, and legal factors, systematically disentangling its drivers is essential, and a consistent monitoring of the process provides the data basis for informed decisions and offers insights to project developers, policymakers, and the public. Such systematic, global monitoring requires the automated separation of the individual activities that characterise the deployment phase at the turbine level, from site investigation and foundation construction to the subsequent installation of the pole, nacelle, and rotor blades. Resolving these stages, in turn, allows construction activities to be precisely related to other relevant conditions, including environmental impacts \citep{Ouro_2024}, logistical bottlenecks, legal regulations and incentives \citep{LIN2026101967, WEI2021110366}, animal behaviour \citep{lai2024endangered}, impacts on the maritime boundary layer due to wake effects \citep{Djath_2026} and near-surface climate \citep{Akhtar2022}, ship interaction, fishery \citep{SZOSTEK2025114555}, and seafood production yields \citep{QU2023106811}.

Due to the vast open space to cover, the integration of different data sources, the necessity of fine grained and detailed temporal and semantic resolution of processes, and data access restrictions at a global scale, the collection of all necessary information to prepare a consolidated global picture is a challenging task. Satellite-based solutions enable global-scale monitoring that is independent of project reports and site accessibility, and provides an independent evaluation and monitoring tool. Remote sensing research has made considerable progress in offshore wind turbine mapping from Earth observation data, providing highly automated spatial localization frameworks of offshore wind turbine sites and open access to location data of offshore wind energy infrastructure facilities. Global-scale products focusing on offshore wind turbines have been provided by \citet{zhang2021gowt} and \citet{hoeser2022deepowt}, both leveraging Sentinel-1 SAR imagery to derive turbine locations and to mark the onset of fully deployed turbines with annual and quarterly temporal precision, respectively. \citet{Paolo2024} also use SAR data to detect persistent offshore infrastructure more broadly and transient objects like vessels, and classify each target by investigating both Sentinel-1 and Sentinel-2 imagery of the identified targets with a deep learning approach, resulting in a binary signal for offshore wind turbines that resolves a turbine as deployed or not deployed at a monthly temporal granularity. Another study by \citet{Zhang2024gowtgeedeep} investigates Sentinel-2 imagery for offshore wind turbine detection and then uses monthly aggregated Sentinel-1 backscatter temporal profiles to provide a binary turbine -- no-turbine signal whenever the backscatter amplitude changes significantly at detected sites.

Further national studies focus on Chinese offshore areas \citep{LIU2026108706, liu2024shandong, wang2024owtchina, Ding2024owtcn, song2025seanson, he2025onoffshore}, which is the world's largest and most dynamic market for offshore wind energy \citep{hoeser2022heightcapa, HOESER2026100451}. The study by \citet{WANG2024explosivegrowth} provides the longest time series to date. Derived from Earth observation data, it provides binary turbine -- no-turbine labels at annual temporal resolution by investigating Sentinel-1 imagery for more recent and Landsat-8 imagery for older turbine constructions, compiling a time series ranging from 2000 to 2022.

On a regional scale, \citet{XU2020110167} provide offshore wind turbine locations and binary turbine -- no-turbine temporal sequences for the North Sea and surrounding waters, and \citet{liu2024shandong} do so for the Chinese province of Shandong. Both studies investigate multispectral data and provide the binary turbine signal on an acquisition-date basis, meaning that the binary signal can be updated as soon as a new acquisition becomes available.

Together these studies show that, on a global scale, temporal granularity is often aggregated to monthly or coarser resolution, while on a regional scale approaches for closer monitoring are proposed, even at an acquisition-date granularity \citep{liu2024shandong, XU2020110167, HOESER2026100451}. However, even when each available acquisition of one or multiple Earth observation missions is used to guarantee a high temporal resolution, communicating offshore wind turbine construction as a binary event largely underestimates the complexity of the deployment process. Combining global-scale offshore wind turbine mapping with a temporal analysis of the SAR signal at the turbine location, by investigating the backscatter intensity values along the range direction, allows more detailed insights into the dynamics occurring at offshore wind turbine locations and makes it possible to differentiate between foundation construction phases and fully deployed turbine phases at a quarterly temporal resolution, as demonstrated by \citet{hoeser2022deepowt} introducing the open data set DeepOWT v1.21.2. By further increasing the temporal resolution to unaggregated single acquisitions, \citet{HOESER2026100451} updated the DeepOWT data set (v3.25.1), which includes over 14 million analysis ready 1D SAR profiles of backscatter intensities in range direction, associated with more than 15 thousand offshore wind infrastructure locations worldwide. With this increased temporal resolution, the semantic resolution had to be increased as well, in order to account for signatures showing vessels at turbine locations prior to the actual deployment phase during site investigations, as well as large construction barges present during deployment, both of which are now detectable and separable when looking into single acquisitions. Thus, the DeepOWT v3.25.1 data set contains labels derived with a rule-based classifier for all 14 million events, as well as a hand-crafted test set to foster benchmarking in subsequent studies.

This study directly addresses one of the research gaps discussed by \citet{HOESER2026100451}, namely the limitation of an expert-knowledge-driven, rule-based approach for event classification in an offshore wind turbine life cycle, and instead investigates deep learning based approaches to further increase the predicted event label quality, with the goal of specifically disentangling and investigating the deployment phase of offshore wind turbines. To our knowledge, the work presented here provides the first systematic investigation of how deep learning can improve dense classification of Sentinel-1 based offshore wind infrastructure time series in order to provide a unified model usable at a global scale. The goal of this study is to advance Earth observation-based offshore wind infrastructure monitoring towards a more nuanced and detailed analysis. This is in line with the emerging trend in offshore wind infrastructure mapping of not only answering the question of where the facilities are, but also addressing what specifically is happening at a facility location and at which point in time \citep{HOESER2026100451}. While going into a detailed localized focus on temporal signals, the aim is to preserve the global applicability of the approach and to provide solutions that match the high-volume data archives that Earth observation provides today, in order to answer research questions at a global scale. To this end, the contributions of this study are as follows:

\begin{itemize}
\item We provide access to an additional 500 hand-labeled, globally distributed offshore wind infrastructure event time series, containing 661,732 labeled events as a basis for deep learning model training.
\item A systematic comparison of ten deep learning model--training variations for dense classification of Sentinel-1 based offshore wind infrastructure time series.
\item A comparison of the model performances and benchmarking with the baseline approach on an open benchmark data set.
\item A complete prediction set of 14,840,637 labels from the best-performing deep learning model and an ensemble variant, with the deployment phases explicitly isolated from the derived labels for offshore wind turbine locations ranging 2016-01-01 to 2025-03-31.
\item A demonstrative spatio-temporal analysis of the deployment phase in the major markets China, the European Union, and the United Kingdom from 2016 to 2025, based on the derived labels.
\end{itemize}

\section{Data and Materials}
\label{sec:data}

\begin{figure}
	\centering
	\includegraphics[width=0.9\linewidth]{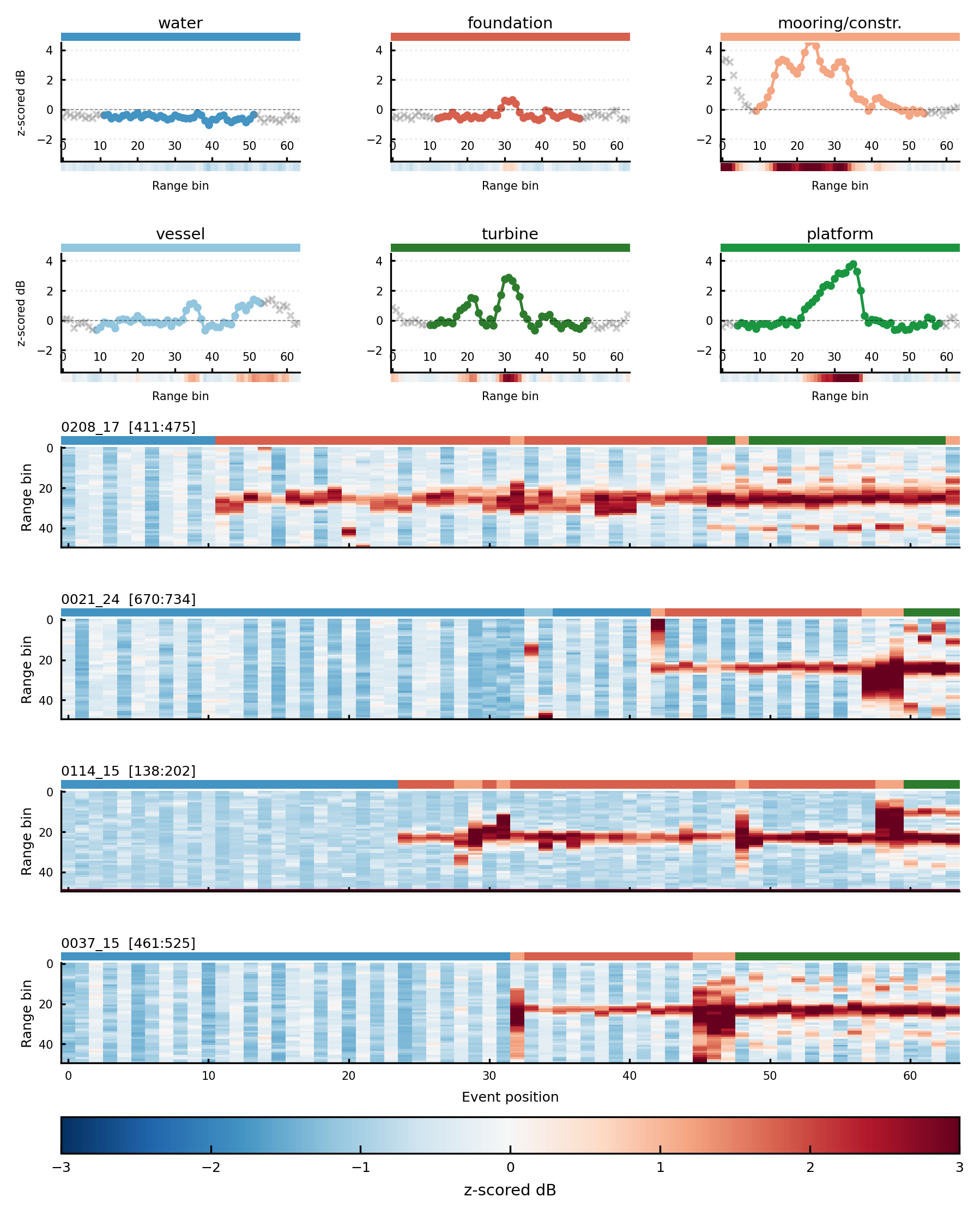}
	\caption{The upper rows show panels of example Sentinel-1 SAR backscatter profiles taken at offshore wind infrastructure locations for each of the six class labels, grey symbols in the profiles indicate the symmetrical padding to 64 bins. The lower four rows show sequences of such profiles ordered by time. The chosen sequences with 64 event profiles are centered around deployment events to provide a visual impression of the investigated temporal signal.}
	\label{fig:profile_overview}
\end{figure}

Figure~\ref{fig:profile_overview} shows examples of single time series events as 1D profiles of Sentinel-1 SAR maximum backscatter intensities at a facility location, and time series of these profiles from the data set investigated in this study. The original data set contains 15,606 time series with 14,840,637 events, each associated with an offshore wind infrastructure location such as a wind turbine or a supporting platform like a substation or meteorological mast. The original data set was curated by \citet{HOESER2026100451} to provide access to dense, global scale offshore wind infrastructure time series derived from Sentinel-1 SAR backscatter intensity values, ranging from 2016-01-01 to 2025-03-31. The temporal resolution of the compiled sequences is not fixed and ranges from 1 day in northern European regions to 12 days in most regions outside Europe. During curation, Sentinel-1 ground range detected (GRD) interferometric wide (IW) VH-polarized (vertical sent - horizontal received) images were cropped around detected offshore infrastructure sites into locally focused 2D patches. These patches were then reduced column-wise by taking the maximum backscatter value to encode amplitude changes along the range direction across each infrastructure site. This reduction is motivated by the strong spatial features related to SAR imaging mechanics, such as the layover effect in range direction for offshore wind turbines, which creates highly discriminative features along the horizontal axis. This profile construction has been applied to all available Sentinel-1 acquisitions in the covered time range, for each turbine location worldwide. By using this approach, over 14 million single events were reduced to compact yet feature-rich representations of the SAR signatures at offshore wind infrastructure locations. For a detailed description we refer to \citep{HOESER2026100451}.

Each event in the data set is assigned to one of the classes \textit{water, vessel, foundation, mooring/construction, turbine,} and \textit{platform}. For a visual impression of the data, the six upper panels in Figure~\ref{fig:profile_overview} show typical backscatter profiles for these six classes. Below, example subsequences of 64 events from entire time series are shown, centered around transition periods of deployment related labels of the time series. The subsequences are taken from the hand-labelled set and are visualized together with the ground truth label of each event displayed on top of each sequence. They illustrate the typical progression from open water with no infrastructure, through the first appearances of vessels conducting site investigations, followed by construction vessels installing foundations. The foundations typically remain for some time until, in a subsequent construction phase, the pole, nacelle, and rotor blades are installed, after which the turbine appears with its characteristic SAR signature with two major peaks, one originating from the turbine center and one from the layover-effect-induced amplitude peak in the range direction towards the sensor \citep{HOESER2026100451}.

\subsection{Data set preparation}

\begin{figure}
	\centering
	\includegraphics[width=\linewidth]{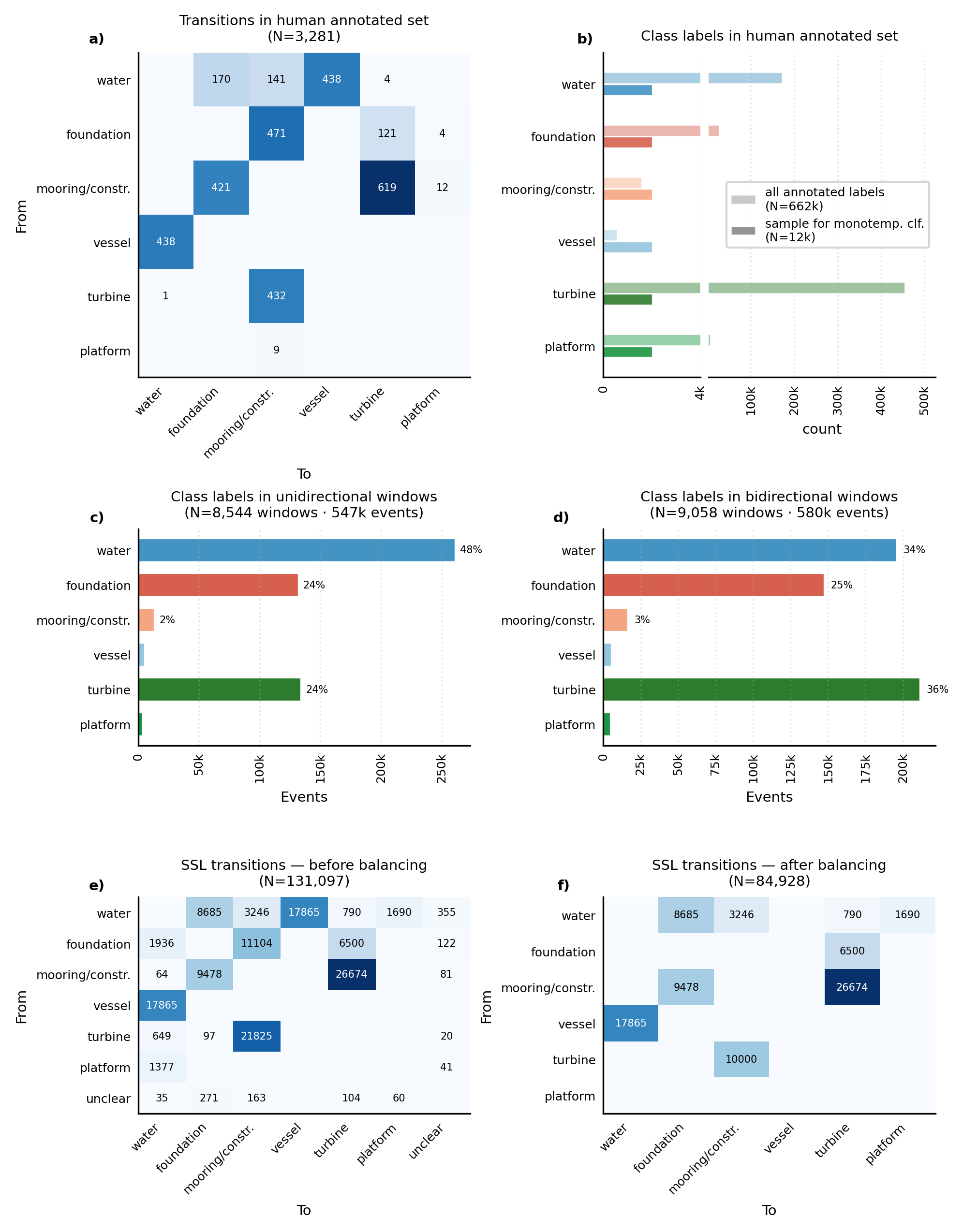}
	\caption{Overview of data set sampling and restructuring to create different deep learning ready data sets from the original data, provided by \citet{HOESER2026100451}, and the additional 500 hand labeled sequences.}
	\label{fig:transitions}
\end{figure}

By labeling an additional 500 time series for this study, we create a basis from which we construct three supervised deep learning training data sets, and, by exploiting the large amount of remaining unlabeled data, construct three additional sets suitable for self-supervised learning (SSL). To prepare the data set for deep learning experimentation, we performed the splits and data sampling shown in Figure~\ref{fig:transitions}. First, we strictly isolated the test events provided in the original data set to serve as a hold-out set for final evaluation and benchmarking. This set contains 553 offshore wind infrastructure facility related time series with 328,657 labeled events. We then randomly selected another 500 time series from the remaining data and hand-annotated each single event in those sequences, creating an additional 661,732 event labels. This hand-annotated set of time series is used to construct deep-learning-ready training data sets for supervised training, each tailored to specific model--training variants.

The event label distribution of the human-annotated set, shown in Figure~\ref{fig:transitions}b), reveals that the raw annotated set contains considerably more turbine, water, foundation, and platform labels than mooring/construction, and vessel labels. This is expected, since the data set focuses on offshore wind infrastructure locations that, throughout the covered time range, are either absent (showing water) or installed (mostly showing a wind turbine). Vessel-related labels, by contrast, are rare events. Since our focus is to explicitly disentangle the events within a turbine's deployment phase, we applied the following strategies to prepare suitable training data sets.

For monotemporal model-training variants, which operate on individual event profiles rather than sequences of event profiles, we randomly undersampled the dominant classes and oversampled the two minority classes vessel and mooring/construction to build a balanced data set with 2,000 examples per class, resulting in 12,000 examples, see Figure~\ref{fig:transitions}b). For multitemporal, model--training variants, single-event balancing is not possible. Instead, we centered the raw annotated set around label transitions using a window size of 64 events, see Figure~\ref{fig:transitions}a). This subsequence cropping is identical to the way sequences have been shown in Figure~\ref{fig:profile_overview}, which can be visited for a visual impression of how a training window can look like. This way, we ensure that the parts of the hand-labeled sequences showing class transitions are included in the training data while reducing otherwise homogeneous sequences. Furthermore, to train the model not only on construction sequences but also on decommissioning sequences, we artificially shifted and flipped windows showing typical construction transitions such as water to foundation and foundation to turbine. These reversed examples increased the training set size by 20\%.

Window placement around transition labels was performed in two variants, resulting in two training data set configurations, unidirectional and bidirectional windows. In unidirectional windows, the transition label is placed at the end of the window, whereas in bidirectional windows it is placed at the center. At each transition position, we sampled one window located exactly at the transition and three additional windows with a random jitter ranging from $-3$ to $-8$ for unidirectional windows, and from $8$ to $16$ for bidirectional windows. During window construction, we added equal padding at the start and end of each sequence and allowed a border overshoot of up to 16 events into the padded region. Finally, to reintroduce fully homogeneous sequences that were avoided in the first step, we added back two training windows per entirely homogeneous sequence, resulting in 170 transition-free windows from 85 transition-free sequences in the raw annotated set. The resulting event label distributions are reported in Figure~\ref{fig:transitions}c) and d). The unidirectional training data set contains 8,544 windows of 64 events each, totaling 547 thousand events, while the bidirectional training data set contains 9,058 windows totaling 580 thousand events. Because the sequences are dominated by water and turbine events, and because platform labels are mostly concentrated in transition-free sequences that are largely discarded, the window-based training data sets show a characteristic class imbalance at the event level, favoring water, foundation, and turbine over vessel, mooring /construction, and platform.

Given the large remaining set (14,553 sequences) of unlabeled and so far unused data, we exploit this potential by preparing training data sets for self supervised learning. To selectively sample specific labels from the remaining 14,553 time series with 13,850,248 events, we leverage the labels predicted by the rule-based classifier of \citet{HOESER2026100451}. For a purely event-based training data set, we sample 40,000 events evenly distributed over the six classes. For window-based sampling, we first extracted all transitions present in the set, as shown in Figure~\ref{fig:transitions}e), and removed transitions we consider implausible, such as deployed turbine to water or any transition involving the \textit{unclear} label from the rule-based classifier. We further removed or undersampled specific transitions (foundation to mooring construction, turbine to mooring construction, and vessel to water) since we keep their opposite transitions, which in these cases often encode the same events. The selected set of transitions is provided in Figure~\ref{fig:transitions}f), to which we apply the same window sampling logic described above for constructing the supervised training sets. This results in two SSL training sets: the unidirectional set contains 255,868 windows and the bidirectional set 219,013 windows. Compared to the supervised sets, this is more than twenty times the size, suitable to contribute data variety during training.

As a final preprocessing step, we clipped the backscatter intensity values $x$ of the SAR profiles to the interval $[-40.0,\,5.0]\,\mathrm{dB}$, where the clipped values are $\tilde{x}$, and computed the empirical mean and standard deviation over all $14$~million SAR profiles consisting of $N = 655{,}272{,}357$ individual backscatter values, as $\mu = \frac{1}{N}\sum_{i=1}^{N} \tilde{x}_i$ and $\sigma = \sqrt{\frac{1}{N}\sum_{i=1}^{N}(\tilde{x}_i - \mu)^2}$. This yields $\mu = -22.67\,\mathrm{dB}$ and $\sigma = 5.98\,\mathrm{dB}$. These statistics are used to standardize each backscatter value when loading the data prior to training and inference, $z = (\tilde{x} - \mu)/\sigma$.

\section{Method}
\label{section:method}

\subsection{Deep learning based event classification}

Deep learning models are well-established choices for time series classification \citep{IsmailFawaz2019, 8742529}. For the specific task of this study, assigning a class label to every event in a sequence, recurrent neural networks \citep{58337, ELMAN1990179} such as Long Short-Term Memory (LSTM) networks \citep{NIPS1996_a4d2f0d2, 10.1162/neco.1997.9.8.1735, 10.1162/089976600300015015} and attention-based transformer models \citep{vaswani2017attention} are chosen in this study. An LSTM produces a so-called hidden state at each step of a sequence, which encodes its input data plus relevant information from previous states. Through its gating mechanism, it can propagate this information across long temporal connections, establishing both long- and short-range relations between the steps it classifies \citep{10.1162/neco.1997.9.8.1735}. In bidirectional layouts, BiLSTM uses both past and future context, otherwise an LSTM would only work directed or causal \citep{650093, GRAVES2005602}. Transformer-based architectures use the self-attention mechanism to establish connections between individual steps of a sequence \citep{vaswani2017attention}. When the attention matrix is left unmasked, so that, similar to the BiLSTM, both past and future context can be taken into account, the attention mechanism connects each step directly to every other step in the sequence \citep{devlin2019bert}. While these architectural building blocks are conceptually promising candidates for the given task, variations in their implementation, combined with the constraints of different application scenarios, require structured experimentation to determine which model is the right choice.

\subsubsection{Experimentation setup}

\begin{figure}
	\centering
	\includegraphics[width=0.9\linewidth]{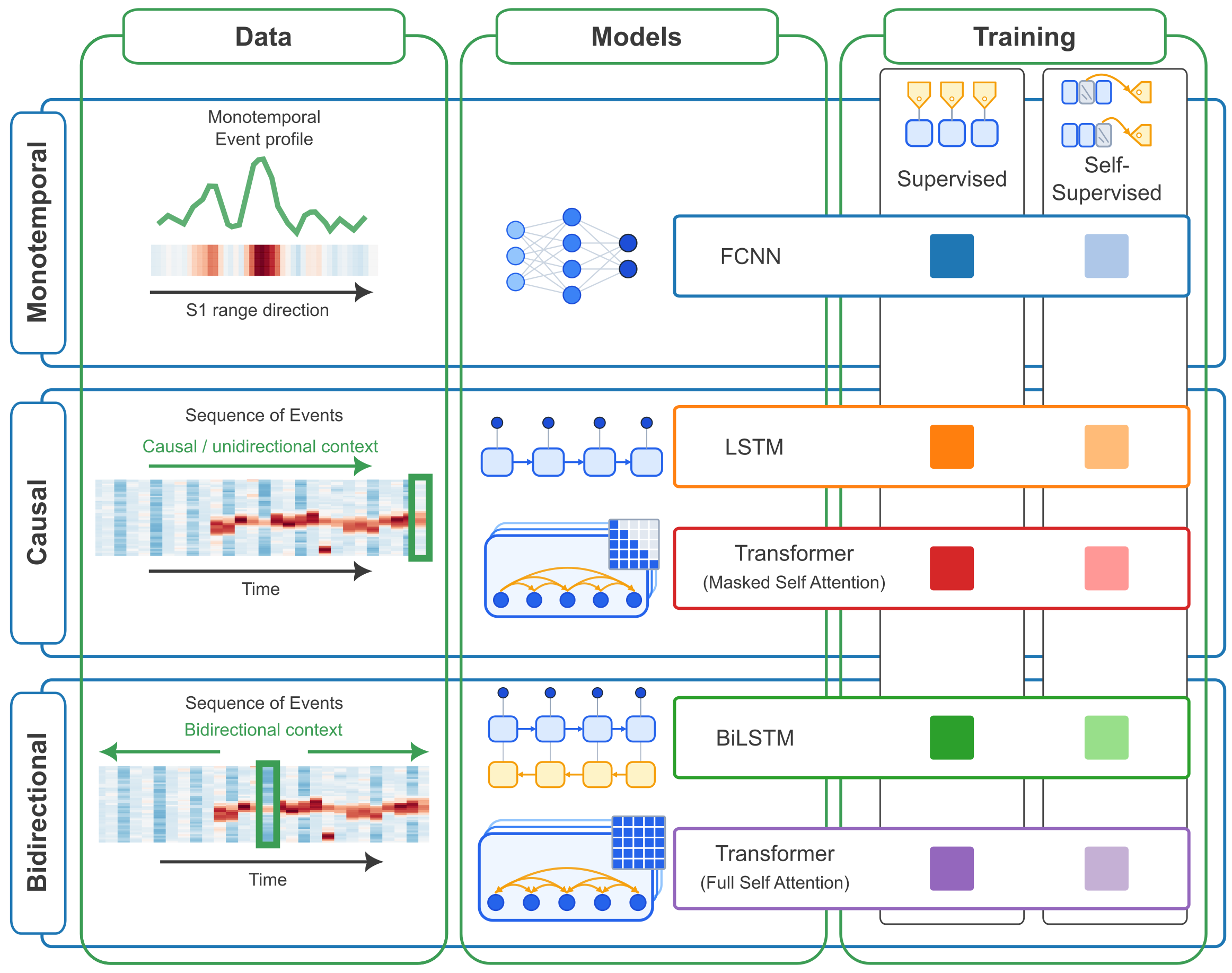}
	\caption{Experiment design for the structured investigation of different data, model and training combinations, centered around how the data is seen, where monotemporal does not contain any sequential information, causal only historic context until the event to be predicted, and bidirectional has access to both, past and future context.}
	\label{fig:exp_design}
\end{figure}

The preceding section on the data prepared for this study already points towards its experimental setup and methodological focus, which is centered around a structured investigation of deep learning model and training variants in order to identify the best-performing models and the most promising deep learning approaches for dense classification of offshore wind infrastructure temporal sequences. To this end, we conduct a set of experiments, structured as depicted in Figure~\ref{fig:exp_design}.

The proposed experimental setup is driven by three major design choices:

\begin{itemize}
\item Compare baseline models without sequential awareness to models with sequential awareness --- monotemporal stream.
\item Train models with access to data \textbf{around} a potential class transition, giving them the capability to observe events before \textbf{and} after potential label transitions --- bidirectional stream, useful for offline inference, such as historic bulk processing of Earth observation archives.
\item Train models with access only to historic data \textbf{before} a potential class transition, located at the latest events in a sequence --- causal stream, useful for online inference in near-real-time or streaming applications of Earth observation data.
\end{itemize}

These design choices are already reflected on the data side through the preparation of the deep-learning-ready data sets. One data set containing event--label pairs without any sequential structure, and two data sets containing unidirectional and bidirectional windows with label transitions placed at the window end or center, have been prepared. Correspondingly, the model side requires architectures that match the three streams. For monotemporal classification, fully connected neural networks are used, for multitemporal data sets, LSTM and transformer-based architectures are chosen. For the bidirectional stream, the standard LSTM model is extended to a BiLSTM model \citep{GRAVES2005602}, whereas for the causal stream a standard LSTM is sufficient. For the transformer-based architectures, both streams, causal and bidirectional, use the same overall architectural layout. The difference is that for the causal stream the attention matrix is masked (masked self-attention), as is common in decoder-only LLM architectures \citep{Radford2018ImprovingLU, NEURIPS2020_1457c0d6}, whereas for the bidirectional stream the attention masking is disabled (full self-attention) so that the model can attend to all events in the sequence directly, as is more common in encoder-only architectures such as BERT \citep{devlin2019bert}.

To investigate whether supervised learning can benefit from self-supervised pretraining, each model variant is trained both with and without self-supervised pretraining (SSL) \citep{6795261, devlin2019bert} in order to measure its effect. Overall, the combination of monotemporal versus multitemporal processing, causal versus bidirectional context for the multitemporal case, the choice between LSTM and transformer architectures, and finally purely supervised training versus self-supervised pretraining followed by supervised fine-tuning results in the ten experiments shown in Figure~\ref{fig:exp_design}.

\subsubsection{Model architectures}

The model architecture design process was initialized by how the deep learning ready data has been prepared, to match model input and data structure. The deep-learning-ready data sets are structured on two levels:

\begin{itemize}
\item For the multitemporal data sets, the events are grouped into sequences of unified window length $L = 64$ events.
\item Each event is a one-dimensional backscatter profile of variable length, which we center to a fixed size of $E = 64$ bins.
\end{itemize}

Backscatter profiles longer than $E$ are cropped symmetrically, while shorter profiles are symmetrically padded to fill the empty bins, see the six example panels in Figure~\ref{fig:profile_overview} for a visual impression. Each event is represented by the z-scored backscatter values $\mathbf{s} \in \mathbb{R}^{E}$ and a binary valid-bin mask $\mathbf{m} \in \{0,1\}^{E}$ that encodes real bins as $1$ and padded bins as $0$. Stacking both along a channel axis $C = 2$, a single event becomes $\mathbf{x} \in \mathbb{R}^{C \times E}$, and a sequence is a stack of $L$ such events, yielding a tensor $\mathbf{X} \in \mathbb{R}^{L \times C \times E}$. The corresponding batched tensor shape is $(B,\, L,\, C,\, E) = (B,\, 64,\, 2,\, 64)$ for multitemporal and $(B,\, C,\, E) = (B,\, 2,\, 64)$ for monotemporal data sets.

These data shapes motivated a shared event encoder architecture, which we reuse as the entry point for all model variants, since the $C$ and $E$ dimensions are present in all data set variants. The event profile encoder is implemented as a three-block 1D convolutional neural network. The blocks use kernel sizes of 5, 3, and 3 with 32, 64, and 128 output channels, respectively, each followed by batch normalization, a ReLU activation, and a residual connection. The network ingests the two input channels, the z-scored backscatter values and the binary valid-bin mask. The mask is reused by the pooling operations so that only real range bins contribute and padded values are ignored. The resulting feature maps are pooled by concatenating a masked average-pooling and a masked max-pooling operation, and the concatenated output is projected to a 128-dimensional embedding vector by a linear layer.

All architecture variants build on this event profile encoder. For the monotemporal stream, a fully connected network is placed directly on the encoder output. It consists of two hidden layers with 256 and 128 neurons, ReLU activations, and dropout ($p = 0.2$), followed by a linear classification head with 6 output classes.

For the multitemporal streams, each event in the window is encoded independently by the event profile encoder, producing a sequence of 128-dimensional embeddings that are then passed to the respective sequence model. The causal LSTM is a two-layer unidirectional LSTM with a hidden size of 256 and dropout ($p = 0.2$), again followed by a final linear classification layer. The bidirectional LSTM (BiLSTM) uses the same configuration but processes the sequence in both directions and concatenates the resulting hidden states, doubling the output size to 512 before the final classification layer with 6 output classes.

The transformer models share a common layout of 4 transformer layers with 8 attention heads each, with the 128-dimensional encoder embeddings projected to a model dimension of 192 by a linear input projection. The feed-forward network in each transformer block expands the model dimension by a factor of 4 to 768 before projecting it back to 192. Positional information of the position of an event within the sequence, is encoded via Rotary Position Embeddings \citep{su2024roformer}. For causal (unidirectional) windows, an upper-triangular mask is applied to the attention matrix so that each event attends only to past events. For bidirectional windows, no masking is applied, enabling full bidirectional attention across all events in the window. All models output per-event class logits over 6 classes via a final linear head.

\subsubsection{Training configurations}

During training, each model is trained independently and end-to-end. While the architectural layout of the event encoder is identical across all models, weights are not shared between models, but optimized together with all other parameters of the respective architecture. In all variants, the main training stage is supervised learning, in which the event class labels from the hand-annotated set provide the training signal. In addition, each architecture has one variant in which self-supervised learning is applied prior to supervised learning, see Figure~\ref{fig:exp_design}.

For supervised training, we use a multi-class cross-entropy loss and the AdamW optimizer \citep{loshchilov2019adamw} with a weight decay of 0.0001 and a cosine annealing learning rate schedule \citep{loshchilov2017sgdr}, starting at 0.001 with a linear warmup over the first 100 steps. We use gradient clipping with a maximum norm of 1.0, applied to the LSTM- and transformer-based architectures to stabilize training. Training is configured to run for a maximum of 40 epochs with a batch size of 128. Early stopping monitors the validation macro F1 and halts training after 25 epochs without an improvement of at least 0.0001.

\begin{figure}
	\centering
	\includegraphics[width=\linewidth]{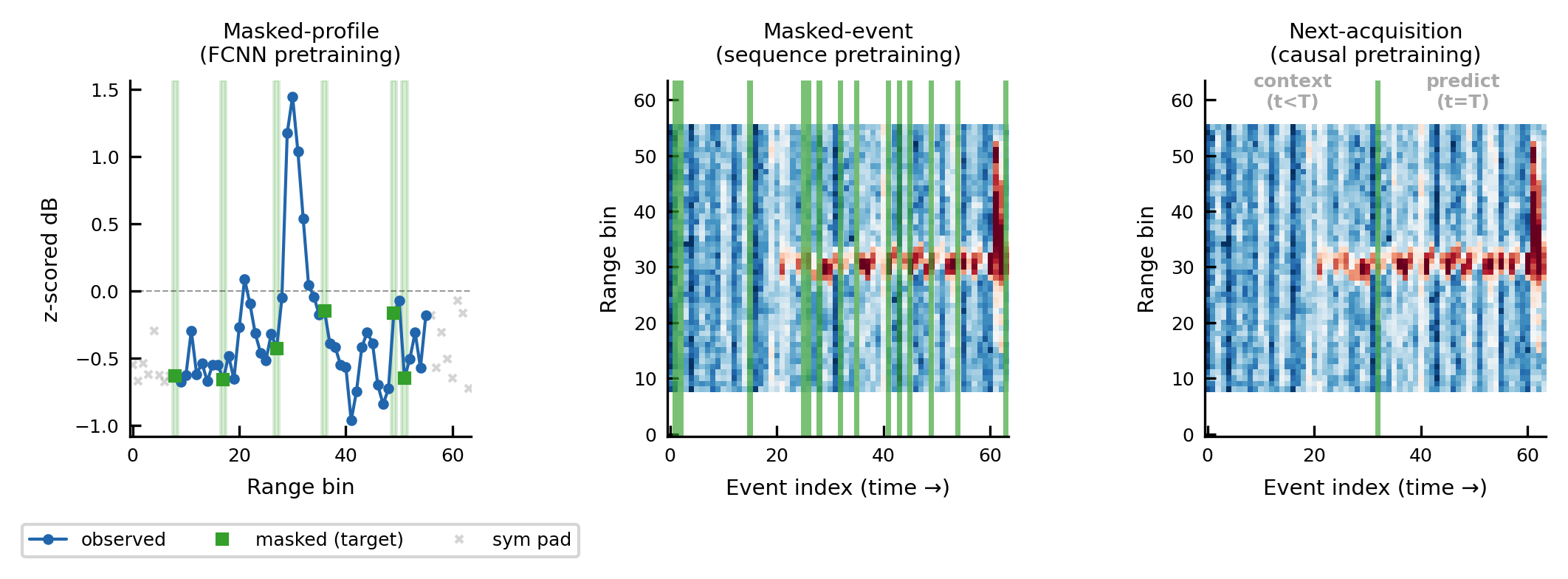}
	\caption{Visualization of the three different self supervised learning data preparation processes. Masking is applied randomly to event profiles (single values) and sequences (entire event profiles) which then have to be reconstructed, whereas for causal models, the pretraining event to be reconstructed is not chosen randomly but is always the next event profile in a sequence of profiles.}
	\label{fig:ssl_masking}
\end{figure}

For variants that begin with self-supervised training prior to supervised training, samples from the prepared SSL training sets are loaded and the mask labels for the reconstruction task are generated on the fly, see Figure~\ref{fig:ssl_masking}. For the FCNN models, which perform monotemporal event-level classification, 15\% of the backscatter intensity values lying within the valid mask (not padded) are masked. Analogously, for bidirectional event sequences, 15\% of the event positions within each sequence are masked out, thus the masks are applied to entire events rather than to individual values within an event. For causal pretraining, no such mask is applied, but the task is rendered as next-prediction to comply with the causal perspective of the architectures to be trained. The model's task is to reconstruct the next event in a sequence from all preceding events in the window, which provides the directed, causal perspective for this pretraining variant.

For all reconstruction tasks, we use a mean squared error reconstruction loss. The pretraining phase runs for 15 epochs without early stopping, using the same training configuration as the supervised stage but with an extended warmup phase of 150 steps. For the subsequent supervised training of self-supervised pretrained models, training starts with a linear 150-step warmup to the peak learning rate of 0.001. To preserve the representations learned during pretraining, we adopt the concepts used by \citet{howard2018}:

\begin{itemize}
\item a discriminative learning rate, which applies different learning rates to different parts of the architecture in order to update the pretrained weights conservatively,
\item and gradual unfreezing, which progressively adds parts of the model to the optimization process, starting with those that have not been pretrained.
\end{itemize}

\begin{figure}
	\centering
	\includegraphics[width=\linewidth]{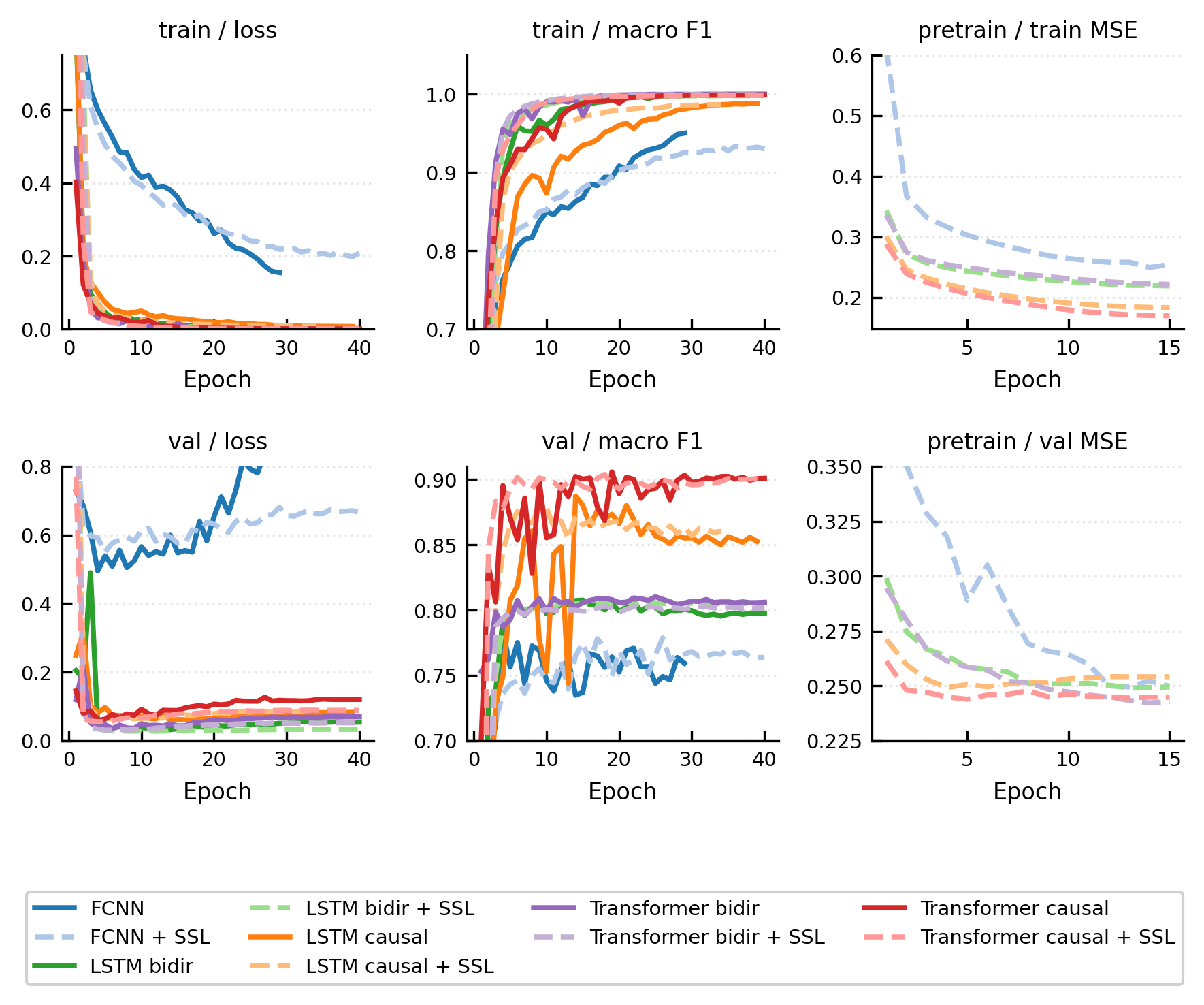}
	\caption{Training curves across all ten model--training variants.}
	\label{fig:training_curves_all}
\end{figure}

During the subsequent supervised training, optimization of the body is disabled over the first 50 steps of the warmup phase, allowing the newly added classification head to adjust its randomly initialized weights and catch up with the SSL-pretrained body. Afterwards, the body's weights are unfrozen (gradual unfreezing). To further protect the pretrained representations, we scale the learning rate of the body's parameters by a factor of 0.1, so that only the classifier head reaches the peak learning rate of 0.001 and consistently operates at a higher learning rate than the pretrained body (discriminative learning rate). All other training configurations remain as described above, including early stopping. Each training data set is divided into a train and validation split with an 80:20 ratio. Figure~\ref{fig:training_curves_all} reports the supervised and SSL losses and the macro F1 used for early stopping, for both the train and validation splits.

\subsubsection{Inference}

All experiments produce trained model checkpoints, which are used to predict event labels on the benchmark test set. Based on these performances, the best-performing model is then selected for a full inference run over the entire data set, covering all 14 million events.

During inference, for the monotemporal FCNN, events are passed through the model independently. For the multitemporal models, windows of 64 events are constructed for each facility sequence with a stride of 1, guaranteeing that every event receives exactly one prediction. For causal models, the last position in each window is kept as the predicted label, and for bidirectional models, the center position is kept. To predict at sequence edges, where a full window cannot be constructed, equal padding is applied by repeating the first event profile to fill the left boundary and, where necessary for bidirectional models, the last event profile to fill the right boundary. For inference we use a batch size of 512, four times the training size.

\subsubsection{Evaluation}

For the task of dense event labeling of offshore wind infrastructure time series, \citet{HOESER2026100451} recommend the area under the collapsed edit similarity at quality threshold curve as a meaningful metric, as it is better aligned with the class-imbalanced events, for which sequence overlap around label transitions matters most, compared to pure edit similarity and micro F1 scores. Edit similarity is based on the Levenshtein distance \citep{levenshtein1966binary}, which measures sequence alignment by counting the edit operations necessary to match a predicted sequence with its target. In the collapsed version, homogeneous parts of a sequence are reduced to a single label that then enters the evaluation. Edit similarity is defined as:

\begin{equation}
\mathrm{EditSim}(S, \hat{S}) = 1 - \frac{d_{\mathrm{Lev}}(S, \hat{S})}{\max(|S|, |\hat{S}|)},
\end{equation}

with the Levenshtein distance $d_{\mathrm{Lev}}$, normalized by $|S|$ and $|\hat{S}|$, the sequence lengths, accounting for sequences of different sizes. With the test set containing $M$ sequences $\{S_1, \dots, S_M\}$, each with an associated prediction $\hat{S}_j$, the corresponding area under the curve is defined as the fraction of these $M$ sequences whose collapsed edit similarity reaches at least a quality threshold $q$, integrated over $q$:

\begin{equation}
\text{AUC}_{\text{EditSim}} =
\int_0^{1}
\frac{\big|\{\, j : \text{EditSim}(S_j, \hat{S}_j) \geq q \,\}\big|}{M}
\; dq .
\end{equation}

A further important metric, reported together with this curve, is the collapsed edit similarity at the 100\% quality level, or perfect match rate, which expresses a model's capability to predict perfectly aligned sequences. To complement these sequence-level metrics and bridge the gap to the validation macro F1 used for early stopping during training, we also report precision, recall, and F1, both macro- and micro-averaged as well as per class, for a thorough evaluation:

\begin{equation}
\mathrm{Precision} = \frac{\mathrm{TP}}{\mathrm{TP} + \mathrm{FP}},
\end{equation}

\begin{equation}
\mathrm{Recall} = \frac{\mathrm{TP}}{\mathrm{TP} + \mathrm{FN}},
\end{equation}

\begin{equation}
\mathrm{F}_1 = 2 \times \frac{\mathrm{Precision} \times \mathrm{Recall}}{\mathrm{Precision} + \mathrm{Recall}}.
\end{equation}

\begin{figure}
	\centering
	\includegraphics[width=\linewidth]{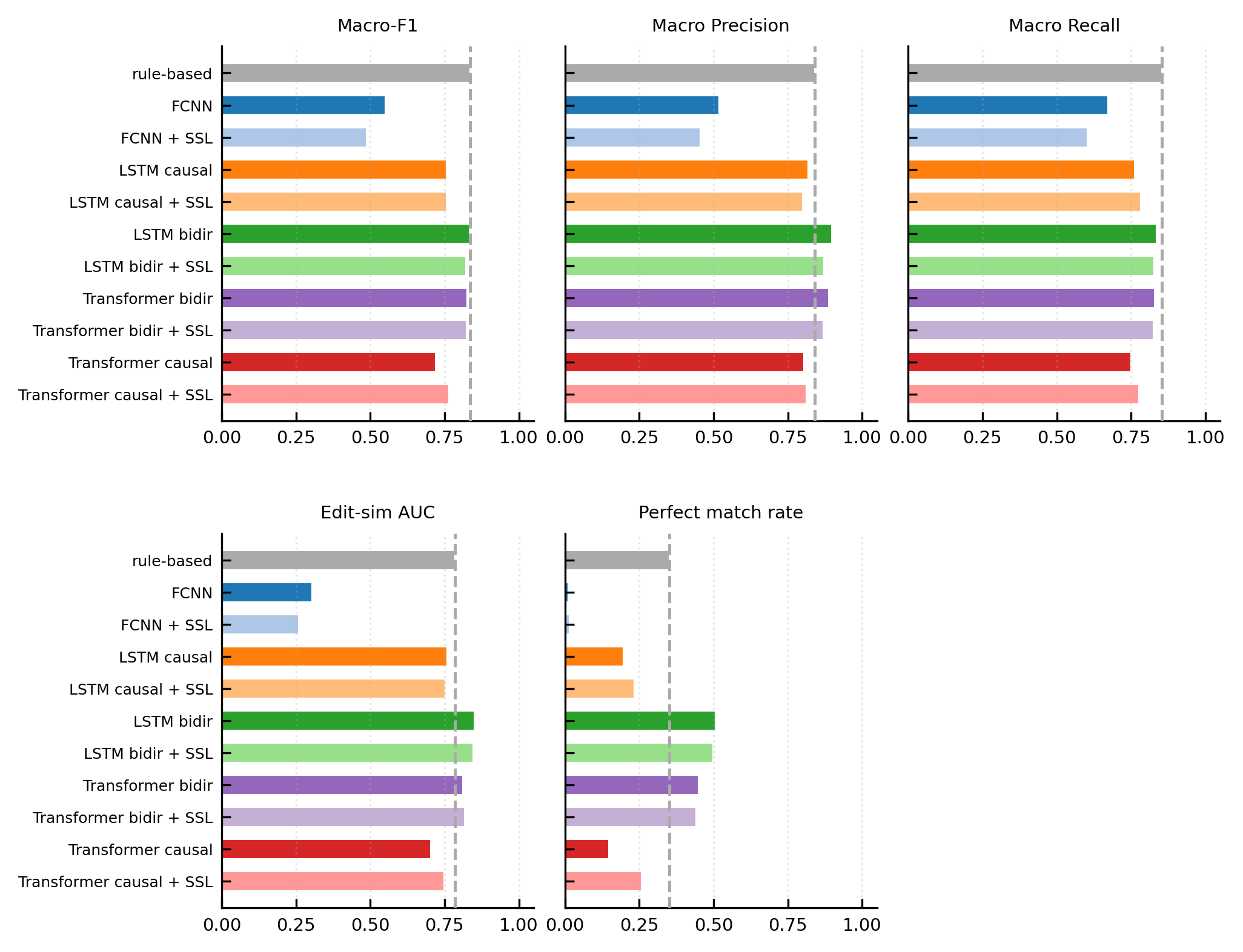}
	\caption{Performance metrics comparison across all model--training variants and comparison with the rule-based baseline classifier performance.}
	\label{fig:metrics_comparison}
\end{figure}

\begin{figure}
	\centering
	\includegraphics[width=\linewidth]{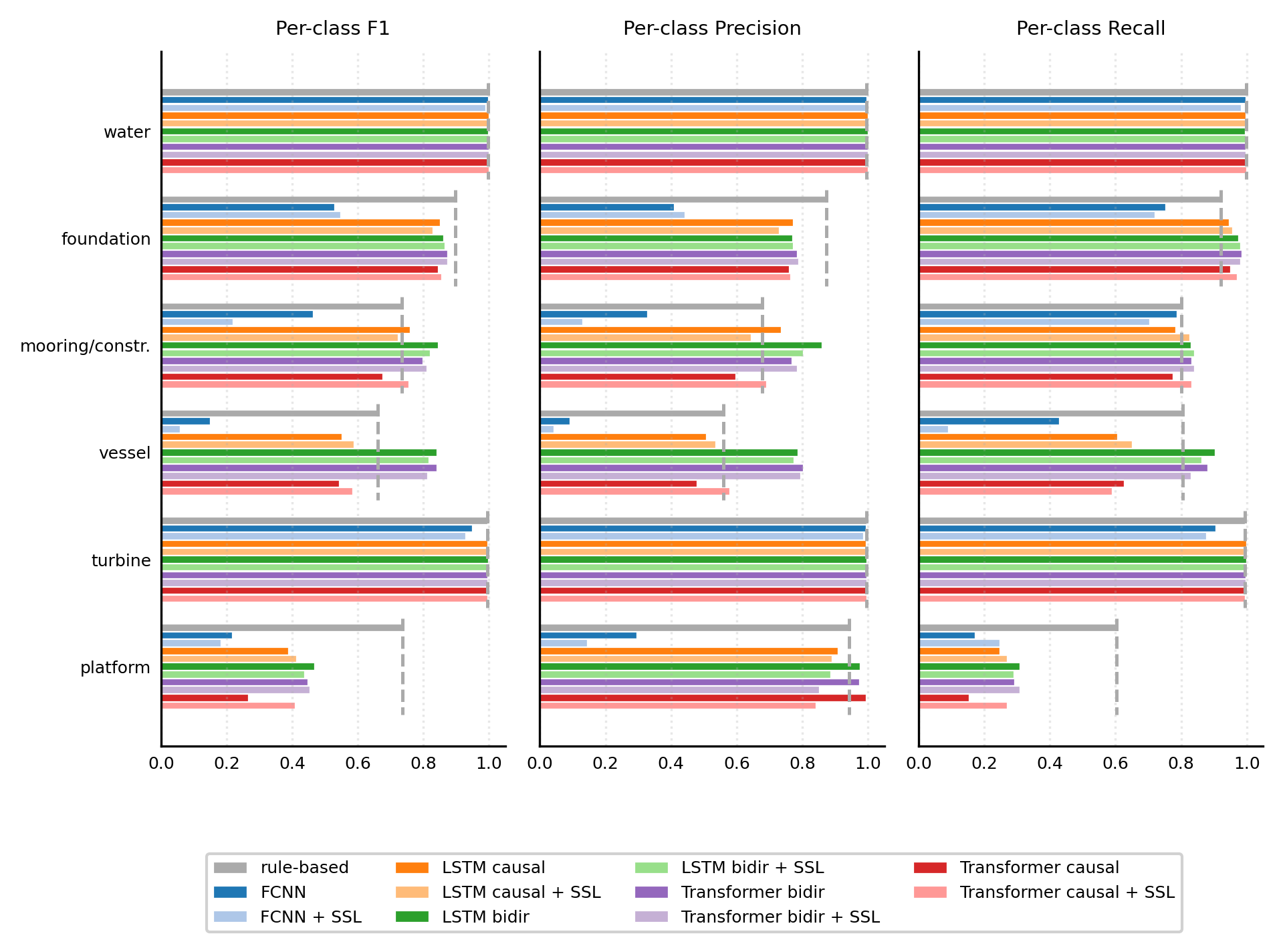}
	\caption{Class-wise performance metrics comparison across all model--training variants and comparison with the rule-based baseline classifier performance.}
	\label{fig:per_class_f1}
\end{figure}

To further investigate the differences between the predicted labels and the baseline labels provided by the rule-based classifier of \citet{HOESER2026100451}, we compare the number of class transitions $T$ in a sequence between the best-performing prediction $\hat{S}$ and the baseline $S$. Investigating the number of class transitions is motivated by the observation that noisy predicted class sequences are associated with false-positive occurrences, whereas the homogeneous sequences resulting from false-negative predictions are less problematic. Following that observation, our goal is to determine which approach minimizes label transitions, or which approach is better aligned with the test data in terms of their number of label transitions. To ground our observation quantitatively, for a sequence $S = (s_1, \dots, s_L)$ of length $L$ we count its label transitions $T$ as the number of positions at which the label changes. For each sequence $j$, we then compute the difference between prediction and target, $\Delta T_j = T(\hat{S}_j) - T(S_j)$, so that $\Delta T_j > 0$ indicates over-segmentation (too many predicted transitions). Furthermore, we compute Spearman's rank correlation $\rho$ between the predicted and target transition counts to express this aspect in a single metric.

\subsubsection{Ensemble compilation}
\label{sec:ensemble_method}

After the transition-based evaluation, we propose a final set of class labels by building an ensemble of the predictions of the best-performing deep learning model of this study (supervised-only BiLSTM) and the rule-based baseline labels of \citet{HOESER2026100451}. The two label sets are combined by selecting, for each sequence, the labels from whichever source has the smaller number of transitions. If the number of transitions is equal, the deep learning based predictions are chosen over the rule-based classifications. After combining the two sets, a full evaluation is performed on this new set as well, which we report on in Section~\ref{sec:ensemble_eval} and compare against the baseline and BiLSTM performances.

\subsection{Regional analysis}

With the improved set of offshore wind infrastructure classification labels, we focus a regional analysis on the deployment phases which can be identified in the data set. First, we filter out non-turbine units, identified as sequences in which a platform label is the first or the last non-transparent label, where water, vessel, mooring/construction, and foundation are treated as transparent labels. Within the remaining set, we define a deployment phase as the transition from open water to a deployed turbine, which may contain intermediate vessel, turbine-foundation, and mooring/construction labels. To suppress isolated false-positive labels that would indicate a false deployment onset, we first identify all sub-sequences of ten or more consecutive water/vessel observations. The deployment end is the first confirmed deployed-turbine observation after the earliest long water/vessel sub-sequence. The deployment start is the first onset label (turbine foundation, mooring/construction, or the deployed turbine itself) occurring after the latest long water/vessel sub-sequence that precedes the deployment end.

Each qualifying unit is then linked, via its spatial coordinates, to an exclusive economic zone (EEZ) provided by the Flanders Marine Institute \citep{VLIZ2024}, and the marked events are dated using the acquisition dates in the original data. For units with both a start and an end label (a fully deployed turbine), we compute the deployment duration in days. For a first regional analysis, we relate the deployment phases to one of the three major offshore wind markets: China, the European Union, and the United Kingdom. We further sort the deployment phases by end date within each market to provide a visual impression of the deployment dynamics, see Figure~\ref{fig:depl_dyn_region}.

For the subregional analysis, we partition the provided EEZs into 8 coastal subregions. This is done by reducing the coastline to a smoothed north–south gradient, split into 8 equal-length segments. The unified EEZs are partitioned by a Voronoi tessellation of 700 random points, each assigned to one of the zones, resulting in 8 automatically derived virtual subregions. We then aggregate the point-located deployment phases per subregion and count all deployment activities from the labeled sequences. Within each deployment subsequence, we count the transitions to foundation and deployed turbine, as well as all labels of class mooring/construction, to get a number of detected deployment activities per deployment phase. We then report the number of deployment activities aggregated by region and monthly bins in Figure~\ref{fig:depl_dyn_cn}, to provide an impression of how the deployment of offshore wind turbines has evolved over time along the Chinese coast, the world's most dynamic and largest hotspot for offshore wind infrastructure.

\section{Results}
\label{section:results}
\subsection{Model performances}

\begin{figure}
	\centering
	\includegraphics[width=\linewidth]{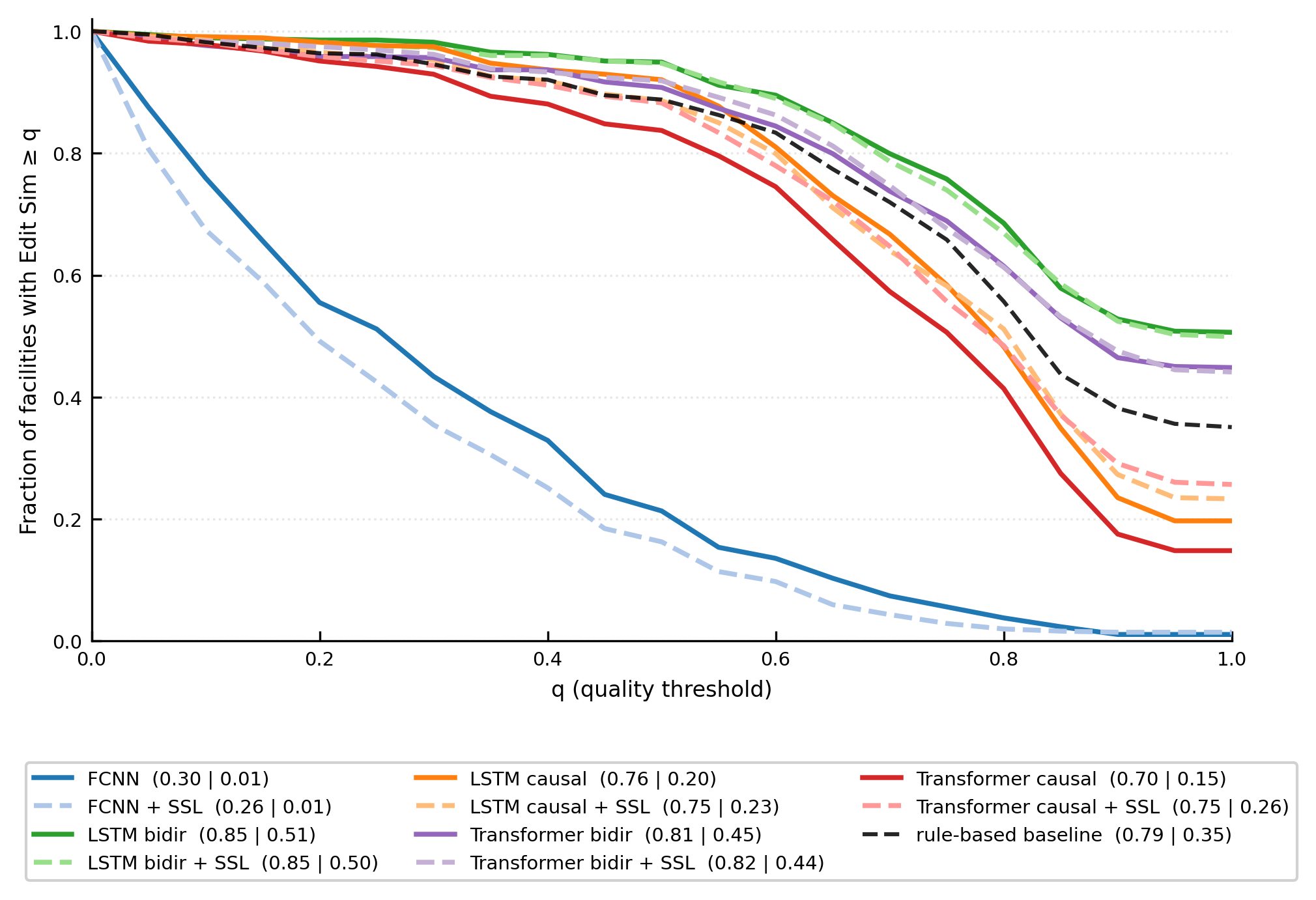}
	\caption{Edit similarity curves for all  model--training variants and comparison with the rule-based baseline classifier. Scores in the legend show ($\text{AUC}_{\text{EditSim}}$ | Perfect Match Rate).}
	\label{fig:edit_sim_auc}
\end{figure}

The experimental setup yields ten models. For each, the following figures report the evaluation results of the checkpoint with the peak $\text{AUC}_{\text{EditSim}}$. Figure~\ref{fig:metrics_comparison} shows that all bidirectional model variants exceed the rule-based baseline classifier, which reaches an $\text{AUC}_{\text{EditSim}}$ of 0.7853, with the supervised BiLSTM performing best at 0.8509. The advantage of bidirectional deep learning based methods is further expressed by the perfect match rate, where the BiLSTM (0.5063) clearly outperforms the rule-based approach (0.3508). The best causal approaches are roughly on par in terms of $\text{AUC}_{\text{EditSim}}$ ($\text{AUC}_{\text{EditSim}}$ | perfect match rate; LSTM causal: 0.7586 | 0.1971, LSTM causal + SSL: 0.7524 | 0.2333, and Transformer causal + SSL: 0.7490 | 0.2568), with the Transformer causal + SSL variant offering the best trade-off once the perfect match rate is accounted for. All causal models nonetheless remain inferior to the rule-based baseline. The monotemporal setups fall furthest behind, with the better of the two, the basic FCNN, reaching only an $\text{AUC}_{\text{EditSim}}$ of 0.3031 and a perfect match rate of 0.0108.

Figure~\ref{fig:per_class_f1} provides more detail by resolving model performance at the class level. A consistent pattern is that the deep learning approaches achieve high performance on the water and deployed turbine classes, with F1 scores close to 1, and are nearly on par with or even surpass the baseline on the ship-related classes vessel and mooring/construction. The foundation class is the only one where the multitemporal models fall behind the baseline, with the BiLSTM reaching 0.8599 against the baseline's 0.8972. The platform class is strongly dominated by the rule-based classifier, which reaches an F1 of 0.7357, here the BiLSTM is again the strongest deep learning model, but reaches only 0.4677.

Comparing not just the summary scores but the full $\text{AUC}_{\text{EditSim}}$ curves shows further details. For the bidirectional variants, additional SSL leaves performance unchanged, and the curves are nearly identical. For the causal variants, the effect of SSL is visible. It is most pronounced for the transformer architecture, which gains 0.05 points in $\text{AUC}_{\text{EditSim}}$ (from 0.70 to 0.75) and improves even more on the perfect match rate, which rises by 0.11 from 0.15 to 0.26. Although these models remain inferior to the baseline, the result shows that SSL can have a substantial impact for specific architecture combinations.

\begin{figure}
	\centering
	\includegraphics[width=0.75\linewidth]{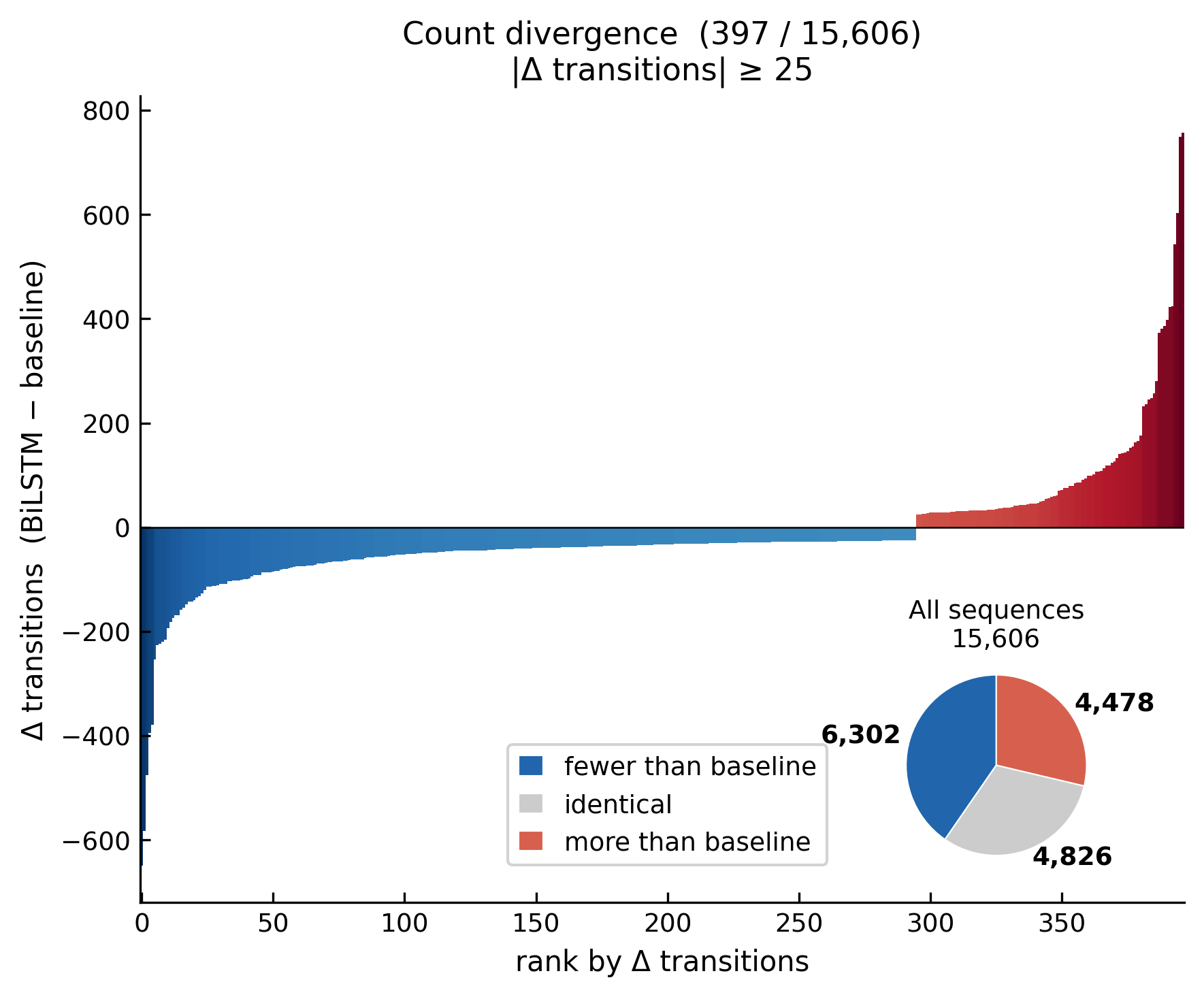}
	\caption{Divergence expressed as the number of label transitions in a sequence between the BiLSTM and rule-based baseline classifier predicted labels across all 15,606 input sequences.}
	\label{fig:label_divergence}
\end{figure}

\begin{figure}
	\centering
	\includegraphics[width=\linewidth]{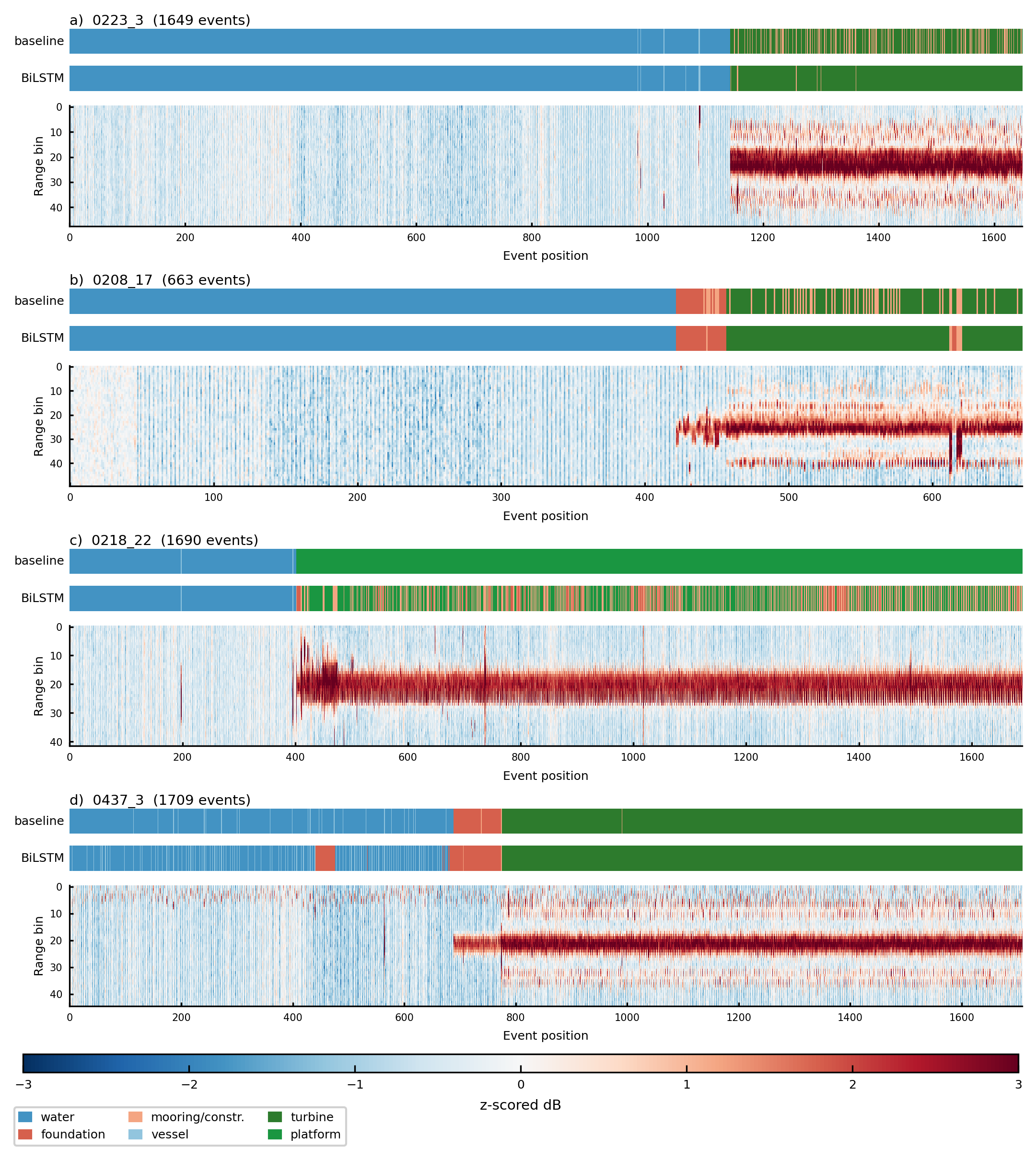}
	\caption{Comparison of four sequences with a+b) noisy labels, predicted by the rule-based baseline classifier and stable labels by the BiLSTM model, and c+d) noisy labels, predicted by the BiLSTM model and stable labels by the rule-based baseline classifier.}
	\label{fig:sequences_dual_labels}
\end{figure}

Example comparisons of prediction results made by the baseline classifier and the BiLSTM model are shown in Figure~\ref{fig:sequences_dual_labels}. Panels a+b) illustrate cases where the BiLSTM produces more homogeneous and more accurate predictions, whereas panels c+d) show cases where the rule-based classifier is more stable. We found that sequences with many transitions tend to indicate misclassifications, and the example sequences support this first impression. Comparing the number of label transitions across all 15,606 sequences of the original, baseline data set (Figure~\ref{fig:label_divergence}), the BiLSTM reduces transitions for 6,302 sequences and increases them for 4,478, while both approaches agree for the remaining 4,826. The BiLSTM thus tends to produce more homogeneous sequences than the rule-based classifier, which is a desirable outcome based on empirical observation that we now want to support with more quantitative arguments.

The preference for models that produce fewer label transitions is supported by a statistical analysis. We correlate the transition counts of each prediction set, the BiLSTM and the rule-based baseline, with the transition counts of the test set. The BiLSTM reaches a Spearman's $\rho$ of 0.89, compared to 0.843 for the rule-based baseline. These statistical findings are in line with our observations that the BiLSTM, leaning towards producing more homogeneous sequences, tends to match better with the real labels. This motivated us to go further into minimizing label transition counts globally for the predictions of the 15,606 sequences. By selectively integrating the derived labels from the BiLSTM model and from the rule-based classifier with the objective to minimize label transitions, we built the ensemble labels as reported earlier in Section~\ref{sec:ensemble_method}.

\begin{figure}
	\centering
	\includegraphics[width=\linewidth]{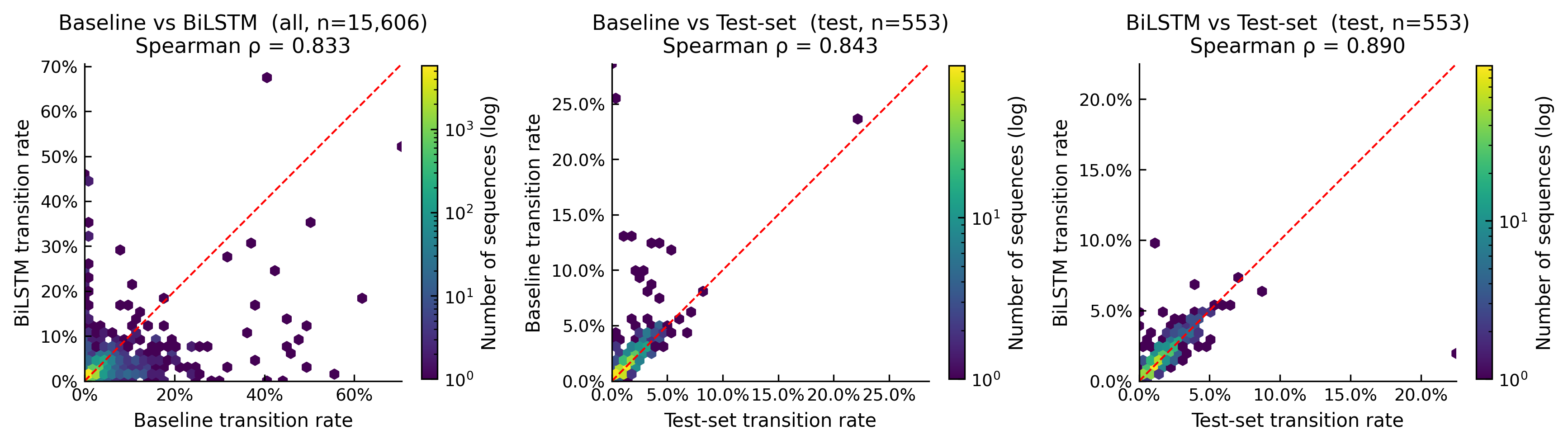}
	\caption{Correlations of transition rates, where transition rates are the number of label transitions in a sequence divided by the sequence length.}
	\label{fig:transition_rates}
\end{figure}

\subsection{Ensemble performance}
\label{sec:ensemble_eval}

\begin{figure}
	\centering
	\includegraphics[width=\linewidth]{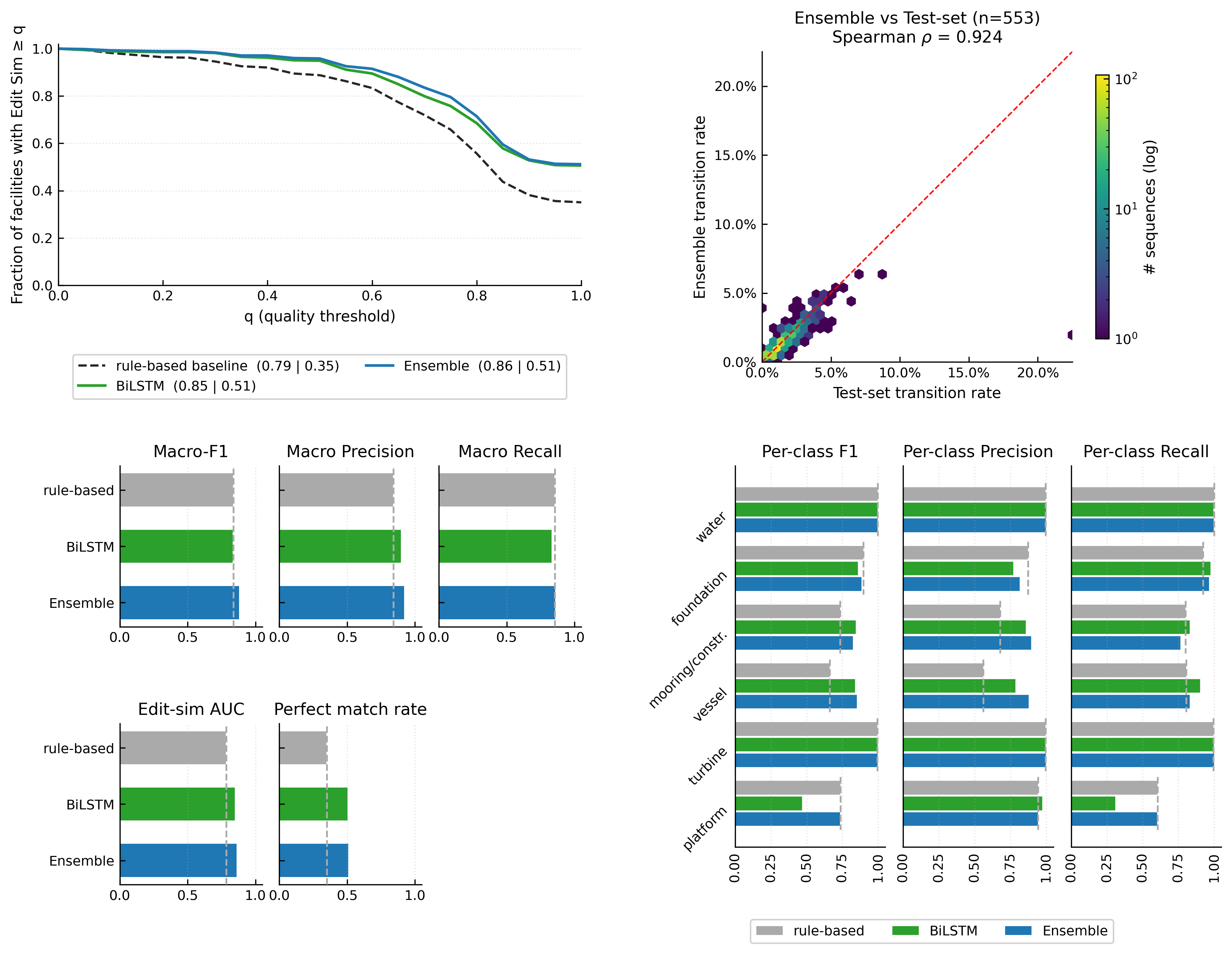}
	\caption{Consolidated performance metric overview to compare improvements of the ensemble results with the two ensemble contributors.}
	\label{fig:ensemble_metrics}
\end{figure}

Figure~\ref{fig:ensemble_metrics} provides an overview of the ensemble labels and compares them to the rule-based baseline and to the best-performing deep learning model, the supervised-only BiLSTM. The ensemble improves $\text{AUC}_{\text{EditSim}}$ only marginally, from 0.85 to 0.86, and leaves the perfect match rate unchanged. Beyond these unchanged summary scores, it has two clearly beneficial effects, which is in line with the assumption that, at the current level of prediction performance, favoring homogeneous sequences with few class transitions over heterogeneous ones with many transitions is the right strategy when combining different approaches.

First, the Spearman $\rho$ between the transition counts of the prediction set and those of the test set rises to 0.924, up from 0.89 (BiLSTM) and 0.843 (baseline). Second, at the F1 per-class level, the ensemble repairs the deep learning models' weaknesses on the platform and foundation classes. This effect is strongest for the platform class, where the class-wise F1 jumps from 0.4677 to 0.7352, matching the baseline's 0.7357. Important to mention is that performance on the other classes does not collapse. The only class with a slight decline is \textit{mooring/constr.}, which falls from 0.8436 (BiLSTM) to 0.8245 (ensemble). From a user perspective, the ensemble is therefore the preferable choice. It also highlights a remaining performance gap between the rule-based classifier and the deep learning approaches, see Section~\ref{sec:discussion} for what these results implicate for future studies.

\subsection{Regional analysis}

\begin{figure}
	\centering
	\includegraphics[width=\linewidth]{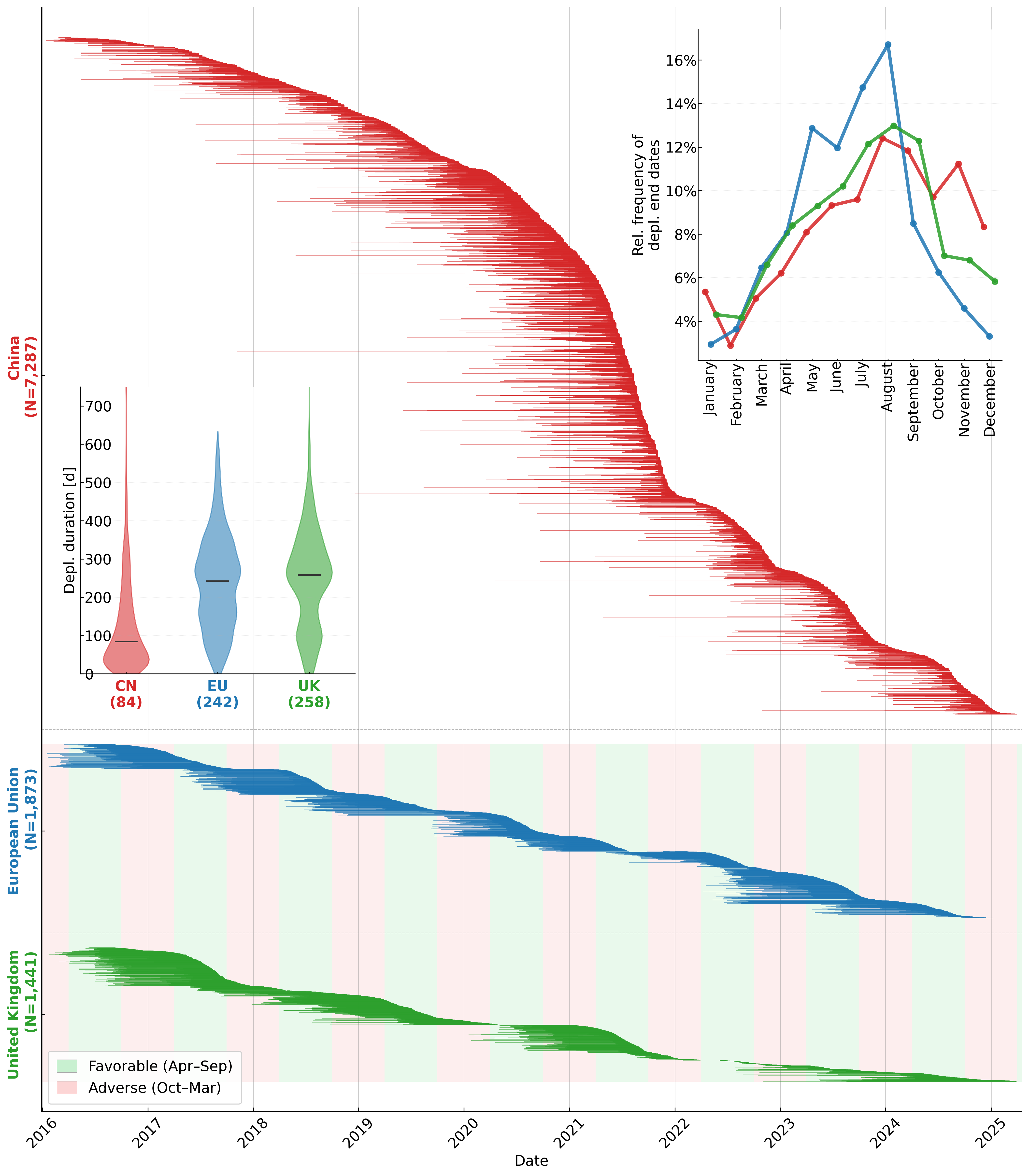}
	\caption{Evolution of deployment phases over time grouped by the three major markets China (CN), European Union (EU) and the United Kingdom (UK). Green and red background color for the EU and UK show periods with favorable (green) and adverse (red) conditions for active offshore wind turbine deployment.}
	\label{fig:depl_dyn_region}
\end{figure}

With the improved event label classification, a detailed analysis of the temporal dynamics of offshore wind turbine deployment becomes more feasible and more reliable. By extracting the deployment phases of turbine-related sequences directly, with start and end dates originating from the Sentinel-1 image acquisition meta data, deployment durations are measured in days. This analysis goes beyond the binary non-turbine--turbine signal which has been the established way of communicating turbine deployment in Earth observation \citep{Paolo2024, LIU2026108706, Zhang2024gowtgeedeep, WANG2024explosivegrowth, zhang2021gowt, liu2024shandong, wang2024owtchina, XU2020110167, Ding2024owtcn}, and it makes a new dimension of insight and analysis from Earth observation data available. Figure~\ref{fig:depl_dyn_region} reports regional differences in deployment durations and patterns, and how they evolved over time, for the three major markets China, the European Union, and the United Kingdom. The median deployment times derived from our improved classification approach are 84~d (CN), 242~d (EU), and 258~d (UK), which are in close agreement with the preceding study, which reported 78~d (CN), 236~d (EU), and 258~d (UK) \citep{HOESER2026100451}.

Beyond the overall deployment duration, Figure~\ref{fig:depl_dyn_region} shows how isolating the deployment phase from the turbine time series exposes regional patterns. The EU and UK both appear to concentrate construction in the favorable window from April to September, when environmental conditions and, in most cases, national regulations allow active work. It can be observed, that the two markets differ in their frequency. The EU follows an annual cycle with steady year-on-year growth in deployed turbines, whereas the UK shows a biennial pattern, marked by longer deployment phases that in some cases span two full years, which aligns with the higher median deployment duration compared to the EU.

The patterns observed in China are entirely different from those in the EU and UK. The deployment data alone do not point unambiguously to a single favorable deployment window that reoccurs every year. However, another pattern reappears, and is getting more pronounced as the Chinese market matures. This pattern is that at the end of each year, deployment activity drops sharply and stays low until around the end of the first quarter of the following year. The year 2021 is especially striking, with a surge of deployments just before year-end. This is driven largely by regulatory and subsidy deadlines expiring at the end of the year, which push projects to commission and grid-connect turbines before the 31-December \citep{LIN2026101967}. That such patterns emerge from the data is a natural demonstration of the approach's applicability and reinforces its value as an independent data source for global offshore wind infrastructure monitoring.

\begin{figure}
	\centering
	\includegraphics[width=\linewidth]{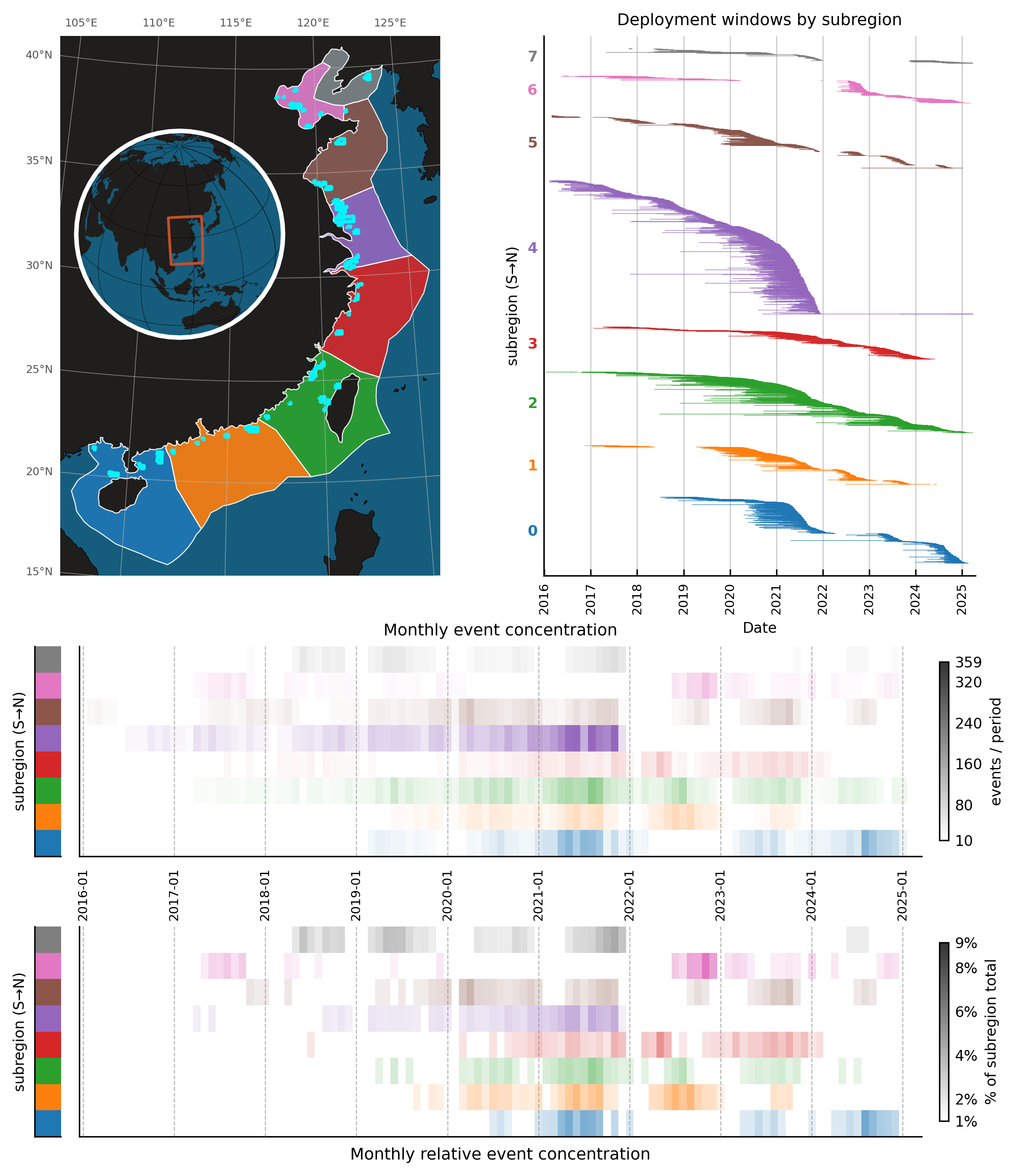}
	\caption{Temporal evolution of the deployment dynamics as derived from the predictions in this study, grouped by automatically generated subregions along the Chinese coast.}
	\label{fig:depl_dyn_cn}
\end{figure}

The subregional analysis of Chinese deployment activity is shown in Figure~\ref{fig:depl_dyn_cn}. An apparent feature is the Jiangsu (Rudong) cluster in subregion 4. Several factors combine to make this the world's densest offshore wind turbine cluster. Most important are the early market entry, see Figure~\ref{fig:depl_dyn_cn}, with turbines built on tidal flats and an (offshore) wind industry supply chain that subsequently clustered in this region \citep{HE201659}, shallow water that extends kilometers offshore as a result of the radial sand ridge system \citep{XING2024107884}, and energy-intensive mainland industry acting as a direct consumer \citep{MIAO2024131417}. Together these provide the geographic, economic, and historical foundation for the region's early development and expansion. The sharp drop and halt in offshore wind turbine deployment, after the end of 2021 raises the question of what drove this significant pattern, observable in the classified time series. The main reason here is not environmental limits but regulatory mechanics. In 2021, the nationwide feed-in tariff guarantee expired for all units grid-connected after 31~December 2021. With no provincial subsidy in Jiangsu left to incentivize deployment, activity appears to have shifted to provinces where such subsidies have been installed, namely Guangdong (subregions 0+1), Shandong (subregions 5+6), and Zhejiang (subregion 3). These subsidies decline in value each year, again with 31~December as the annual deadline \citep{LIN2026101967}, which largely explains the recurring end-of-year deployment surges across these provinces after 2021.

\section{Discussion}
\label{sec:discussion}

\subsection{Bidirectional model variants}
The structured experimental setup in this study provides a new best performing model for dense labeling of offshore wind energy infrastructure time series, accompanied by several insights for further improving model performance. The results show that architectures designed to process sequential data are superior to monotemporal models \citep{RUWURM2020421}, which do not contribute meaningfully to solving the given task. The proposed bidirectional approaches all outperform the benchmark metrics of the rule-based classifier by \citet{HOESER2026100451}, with the BiLSTM leading in performance metrics, demonstrating that deep learning models can encode the rules captured by the rule-based classifier and even more or different ones. Nevertheless, the successful integration of the labels from the rule-based classifier and the labels predicted by the BiLSTM into a result-improving ensemble points to a gap between what the deep learning model has learned and what the rule-based classifier was designed to capture. We argue, that if this were not the case, and had the model learned all details the rule-based classifier encodes, their ensemble would not improve the results. Thus, the gap indicates room to grow for the BiLSTM model, provided the right training signal is supplied.

The direct answer to provide this signal is to increase the amount of labeled training data. Before turning to this straightforward solution, the experimental setup raises another question: if more data is a potential solution, why did self-supervised pretraining not improve performance? Since SSL does not train the task itself but instead uses unlabeled sequences to learn a better representation of the input \citep{6795261, tseng2024lightweight}, our finding is that in the bidirectional case the labeled data alone already allows the model to learn a good representation, and that the limiting factor is instead the discriminative signal the supervised labels provide. The similar performance across all models of the bidirectional stream supports this interpretation, which was also observed by \citep{RUWURM2020421}. If the quality of the learned representation were a bottleneck, SSL variants should consistently outperform their counterparts with no self-supervised pretraining. Since this is not the case, our interpretation concerning bidirectional model improvement is to provide more labeled training data.

Instead of scaling labeled data uniformly, the experimental results allow the residual knowledge of the rule-based classifier to be leveraged directly and render the idea of adding data as disagreement-based active learning \citep{10.1145/130385.130417}. Selecting from the ensemble those sequences characterised by a high number of label transitions between the BiLSTM and the baseline flags the cases in which the two disagree, and thus the cases the deep learning model has not yet captured. These sequences offer training windows with high impact. An additional way is to add examples for such classes which are underrepresented already at the raw data level. The platform class would be one such example, and using recent global-scale mapping of offshore infrastructure such as \cite{Paolo2024} and \cite{Spanier15062026} is a promising starting point to semi-automatically enrich the current raw data with platform sequences.

One counterargument can be made for the latter point, since its practical weight is limited. While improving the discriminative power of the model for platform targets, it contributes little to the offshore wind turbine focus the current application has, and going forward in this direction opens the question of whether a broader model for general persistent marine infrastructure monitoring is a logical conclusion.

\subsection{Unidirectional model variants}

While we discussed that SSL did not help to improve models from the bidirectional stream, this is different for the unidirectional stream. Here, SSL boosts performance in both models, with the Transformer model experiencing the biggest gain. One way to read this is that the transformer model leverages the unlabeled data to learn the temporal representation of the data, whereas the LSTM model as a recurrent model brings this understanding already more strongly encoded in its architectural layout \citep{dosovitskiy2021vit}, thus the gains from SSL are less pronounced for the LSTM model. Increasing labeled training data would also help here. However, the first question to answer for the causal model is whether the current ceiling can be increased when scaling SSL, which is cheaper and whose training objective of next-label-prediction aligns well with the final task.

The reason to pursue improvements for the causal stream, even when the bidirectional stream provides significantly more capable models, is motivated by an application perspective. High performing bidirectional models are the best choice for offline construction of historical data sets and for less time-critical products such as annual reports. A near-real-time service that tracks deployment, operational phase and general activity across offshore wind turbine sites worldwide, independently of terrestrial sensors and networks, could provide situational awareness of critical infrastructure as soon as new acquisitions are available. Such a service demands both precision and timeliness, thus requiring reliable model performance, but from causal models which work unidirectionally and not from bidirectional models since future context is not available in these applications. Currently, causal models do not match the rule-based baseline, so improvements are needed. Focusing solely on the bidirectional stream would lead to further refined data sets, but at the expense of time critical applications.

\subsection{Known limitations}

While the regional deployment-phase analysis shows that meaningful patterns can be derived directly from the predicted labels, and that practical use is therefore possible in general, the deep learning models introduce new challenges that we observed and want to disclose openly as known limitations. During the spatio-temporal analysis we encountered misclassifications of the BiLSTM model that are temporally implausible. Temporal plausibility is a core property the rule-based classifier is explicitly designed to respect \citep{HOESER2026100451}, but the deep learning model failed to learn some of the long range temporal patterns. A typical example is a brief \textit{foundation} phase surrounded by sequences of open water early in a sequence. Although such a configuration is not strictly impossible, the specific cases we observed are misclassifications. They matter in practice, and they are the reason we had to introduce a guarding step when extracting the deployment phases from the predicted labels, where we first isolate long, continuous water subsequences and use them to anchor the detection of the deployment-phase start event.

We attribute this behaviour to the limited temporal context of $L = 64$ that the model sees during training and inference. This context might be too short for the model to learn the global pattern that normally rules out an early \textit{foundation} appearance when a deployment phase occurs later in the sequence, so that the global perspective which would let such a prediction appear implausible is missing. A straightforward solution is to increase the temporal window size. However, this is not straightforward and introduces new pitfalls due to the structure of the data, since the sequences range from roughly 300 to 1700 events while covering the same calendar period. This means that a window defined by a fixed number of observations corresponds to a variable amount of real time. Resolving this, whether by adapting the window to the local acquisition density or by moving to an architecture able to ingest entire variable-length sequences, would substantially change the structure of the required training data, which therefore needs to be addressed soon to avoid wasting costly labeling efforts.

\section{Conclusion}
\label{section:conclusion}

Data from ESA's Sentinel-1 Synthetic Aperture Radar (SAR) mission enables the global monitoring of offshore wind infrastructure. Recent developments towards the analysis of high density time series at infrastructure locations, to investigate events and phases during the infrastructure life cycle, leverage the full potential of these big data archives at an unaggregated, single acquisition level. These fine grained investigations of the temporal dynamics of the local SAR signal need algorithms which can automatically classify major events to process the high-volume, Earth observation data archives at a global scale. In this study, we addressed this requirement by replacing an expert-driven, rule-based event classifier with deep learning models trained for the dense classification of Sentinel-1 based offshore wind infrastructure time series, and by evaluating them in a structured comparison of ten model--training variants.

The results show clearly that architectures with sequential awareness are the most capable for this task, while monotemporal models do not contribute meaningfully. Among those, we further compare causal models (next event prediction with historic context for online inference applications) and bidirectional models (event prediction with historic and future context for offline inference applications). All bidirectional variants exceed the rule-based baseline and the causal variants, with the supervised BiLSTM performing best. For bidirectional models, self-supervised pretraining (SSL) did not improve their performance. We argue, that the labeled data, used in subsequent supervised training, is already providing a sufficient representation. Nevertheless, SSL clearly benefited the causal stream, especially for transformer based models, where its next-label-prediction objective is well aligned with the supervised task and provides meaningful representations for the sequential characteristics of the data. Improvements of causal models are a future research direction, since experiments currently show a potential for these models to improve performance. However, their current performance is below the rule-based classifier.

Using the best performing model of the bidirectional experiment stream, the BiLSTM variant predicts all 15,606 time series in the data set with 14,840,637 events. Combining the BiLSTM predictions with the rule-based labels in a label-transition-minimising ensemble further improved sequence consistency and recovered the performance on under-represented classes, while at the same time pointing to a gap between what the deep learning model has learned and what the rule-based classifier was designed to capture. This allows for future targeted improvements of the labeled data, since the ensemble sequences which reject the deep learning predictions are hard examples that contain signals to further learn from and close this gap.

With these results it was possible to isolate the deployment phase of each single offshore wind turbines at a global scale and substantially improve the semantic granularity of the deployment process, which so far communicated offshore wind turbine deployment as a binary turbine / no-turbine signal. The subsequent regional analysis demonstrates the practical value of this finer resolution, by reporting the deployment durations measured in days, region-specific construction windows, and showing, that repeating, regulation-driven deployment surges emerge directly from the predicted labels.

Overall, this first structured exploration of deep learning model--training variants and benchmarking against a rule-based classifier provides an important overview of the capabilities and limitations of deep learning models in offshore wind infrastructure monitoring from Earth observation data. Future improvements, especially for applications which operate in an online prediction domain and rely on a causal context to make predictions on the latest acquisitions, like scenarios where a near-real-time situation picture of critical infrastructure is of interest, can directly build on this study to improve these challenging but valuable prediction performances. The training data, predicted labels and deployment duration phases are made publicly available to further foster not only deep learning experimentation building on this study, but also offshore wind infrastructure related research and analysis.

\section*{Data Availability}
The data set is publicly available on Zenodo at 
\href{https://doi.org/10.5281/zenodo.5933966}{https://doi.org/10.5281/zenodo.5933966}.

\section*{Abbreviations}

\begin{tabular*}{\textwidth}{r@{\hspace{0.8em}}l@{\extracolsep{\fill}}}
AUC & Area Under the Curve \\
BERT & Bidirectional Encoder Representations from Transformers \\
BiLSTM & Bidirectional Long Short-Term Memory \\
CNN & Convolutional Neural Network \\
EEZ & Exclusive Economic Zone \\
ESA & European Space Agency \\
FCNN & Fully Connected Neural Network \\
FN & False Negative \\
FP & False Positive \\
GRD & Ground Range Detected \\
GW & Gigawatt \\
IW & Interferometric Wide \\
LLM & Large Language Model \\
LSTM & Long Short-Term Memory \\
MW & Megawatt \\
OWT & Offshore Wind Turbine \\
ReLU & Rectified Linear Unit \\
SAR & Synthetic Aperture Radar \\
SSL & Self-Supervised Learning \\
TP & True Positive \\
VH & Vertical transmit, Horizontal receive \\
\end{tabular*}

\section*{Author contributions}
CRediT: Conceptualization TH;
Data curation TH;
Formal analysis TH;
Methodology TH;
Software TH;
Supervision CK;
Validation TH;
Visualization TH;
Writing – original draft TH;
Writing – review and editing TH, FB, and CK.

\section*{Disclosure statement}
No potential conflict of interest was reported by the authors.

\section*{Acknowledgements}
The authors gratefully acknowledge the Copernicus program of the European Space Agency (ESA) for providing free access to Sentinel-1 data. We also thank DLR’s terrabyte team for their dedicated efforts in platform operations, which enabled the deep learning experiments and large-scale inference.

\section*{Funding}
This research did not receive any specific grant from funding agencies in the public, commercial, or not-for-profit sectors.

\section*{AI disclosure statement}
We acknowledge the use of DeepL, which includes a generative AI for English language translation and editing. All AI-generated text suggestions have undergone rigorous revision by the authors.

\bibliographystyle{abbrvnat}
\bibliography{references}

\end{document}